\documentclass{article}

\usepackage[english]{babel}
\usepackage[letterpaper,top=2cm,bottom=2cm,left=3cm,right=3cm,marginparwidth=1.75cm]{geometry}
\usepackage{amsmath,amssymb}
\usepackage{graphicx}
\usepackage[colorlinks=true, allcolors=blue]{hyperref}
\usepackage{cleveref}
\usepackage{mathtools}
\usepackage{tikz}
\usetikzlibrary{arrows.meta} 
\usepackage{caption}
\usepackage{subcaption}
\usepackage{algorithm}
\usepackage{algpseudocode}

\newcommand{\x}{\mathbf{x}}
\newcommand{\X}{\mathbf{X}}
\newcommand{\y}{\mathbf{y}}
\newcommand{\Y}{\mathbf{Y}}
\newcommand{\bmu}{{\boldsymbol{\mu}}}
\newcommand{\bnu}{{\boldsymbol{\nu}}}
\newcommand{\bphi}{{\boldsymbol{\phi}}}
\newcommand{\beps}{{\boldsymbol{\epsilon}}}
\newcommand{\bSigma}{{\boldsymbol{\Sigma}}}
\newcommand{\btheta}{{\boldsymbol{\theta}}}
\newcommand{\bxi}{{\boldsymbol{\xi}}}
\newcommand{\z}{\mathbf{z}}
\newcommand{\Z}{\mathbf{Z}}
\newcommand{\N}{\mathcal{N}}
\newcommand{\Loss}{\mathcal{L}}
\newcommand{\E}{\mathbb{E}}
\newcommand{\DklP}[2]{\ensuremath{\mathcal{D}_{\text{KL}}\left(#1||#2\right)}}
\newcommand{\Dkl}{\mathcal{D}_{\text{KL}}}
\newcommand{\Var}{\mathrm{Var}}
\newcommand{\Grad}{\ensuremath{\boldsymbol{\nabla}_{\!\!\btheta}\,}}

\newcommand{\helge}[1]{\textcolor{blue}{(Helge: #1)}}

\newcommand{\remove}[1]{}

\title{Lecture Notes in Probabilistic Diffusion Models}
\author{Inga Str{\"u}mke\thanks{\href{mailto:inga.strumke@ntnu.no}{inga.strumke@ntnu.no} }\  ~and Helge Langseth\\ 
Norwegian University of Science and Technology}
\date{}

\begin{document}
\maketitle

\begin{abstract}
 Diffusion models are loosely modelled based on non-equilibrium thermodynamics, where \textit{diffusion} refers to particles flowing from high-concentration regions towards low-concentration regions. In statistics, the meaning is quite similar, namely the process of transforming a complex distribution $p_{\text{complex}}$ on $\mathbb{R}^d$ to a simple distribution $p_{\text{prior}}$ on the same domain.
 This constitutes a Markov chain 
 of diffusion steps of slowly adding random noise to data, followed by a reverse diffusion process in which the data is reconstructed from the noise. The diffusion model learns the data manifold to which the original and thus the reconstructed data samples belong, by training on a large number of data points. While the diffusion process pushes a data sample off the data manifold, the reverse process finds a trajectory back to the data manifold. Diffusion models have -- unlike variational autoencoder and flow models -- latent variables with the same dimensionality as the original data, and they are currently\footnote{At the time of writing, 2023.} outperforming other approaches -- including Generative Adversarial Networks (GANs) -- to modelling the distribution of, e.g., natural images.
\end{abstract}

\section{Introduction}
This document aims at being a coherent description of the mathematical foundation relevant for diffusion models. The body of literature in this area is growing very quickly, but the underlying mathematics of the diffusion process remains largely unchanged. In this document we give a self-contained presentation of this foundation, using coherent notation. We will whenever possible avoid discussing issues related to implementation, and rather focus on the fundamental properties of the diffusion models. This document was initially prepared as lecture notes for the course ``IT3030: Deep Learning'' at The Norwegian University of Science and Technology. 


\section{The diffusion process}
\subsection{Forward diffusion}
\begin{figure}
    \centering
    \includegraphics[width=0.95\textwidth]{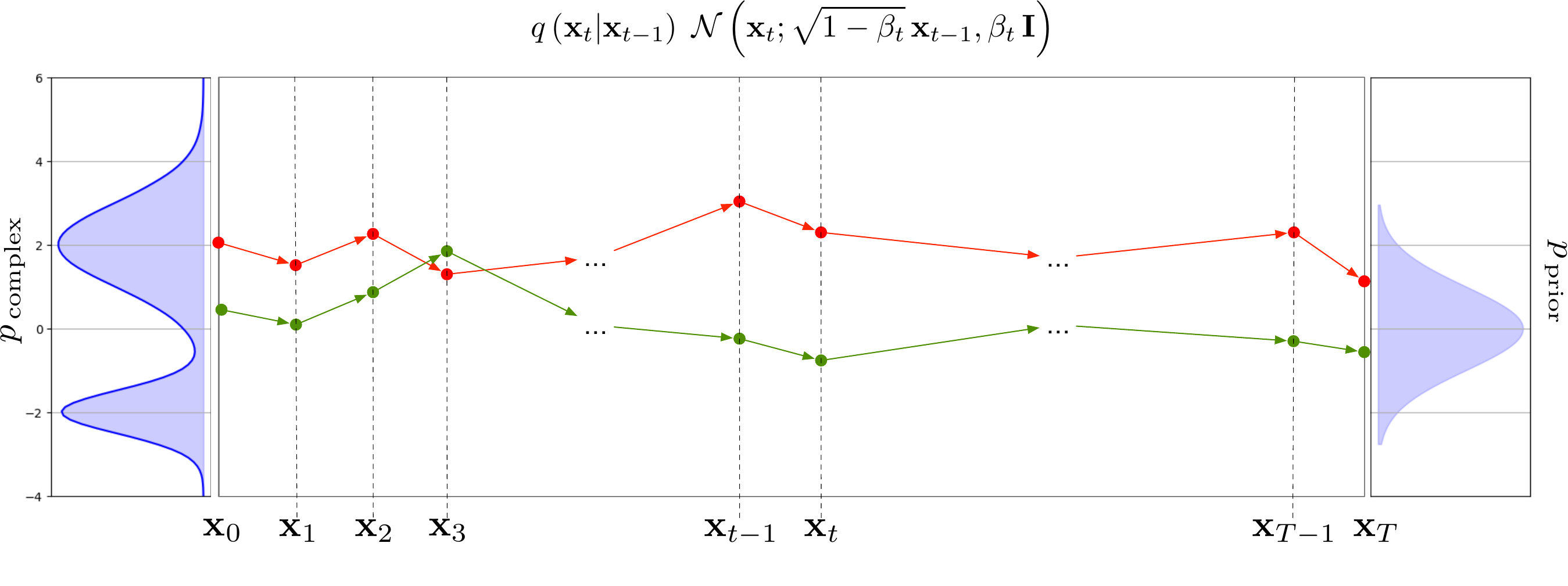}
    \caption{
    \label{fig:diffusion-over-time}
    The forward diffusion process from $p_{\text{complex}}$ to $p_{\text{prior}}$.
    }
\end{figure}

Assume that we have a distribution that we can draw data samples ${\x}_0$ from. The subscript $0$ indicates that this is an original data sample (for instance an image) without any noise added. In the \textit{forward diffusion process}, noise is gradually added to the sample in $T$ steps, generating increasingly noisy samples ${\x}_1, {\x}_2, \dots, {\x}_T$, with ${\x}_T \sim p_{\text{prior}}$ (meaning that the $\x_T$ - samples  follow a predefined distribution $p_\text{prior}$ for sufficiently large $T$). 
The noising procedure must be scheduled to add noise (``destroy'' the data sample) at the right pace. To this end, the variance $\beta$ of the added noise increases following a schedule, i.e.\ the diffusion steps are parameterised by a \textit{variance schedule} $\left\{ \beta_t  \right\}_{t=1}^{T}$. The data distribution is gradually converted into another distribution by repeatedly applying a Markov diffusion kernel $K$, i.e.\  the data sample ${\x}_t$ at step $t$ is generated from ${\x}_{t-1}$ using
\begin{equation} \label{eq:diffusion_kernel}
    q({\x}_t | {\x}_{t-1}) = K\left({\x}_t | \x_{t-1}; \beta_t \right)\,,
\end{equation}
where $\beta_t$ is the diffusion rate.
This makes it clear that the process is Markovian, as each step depends only on the immediately preceding sample. 
The joint probability of the entire process from the original data sample $\x_0$ to the final sample at step $T$ can be written as
\begin{equation} \label{eq:noising-process-q}
    q(\x_{1:T}|\x_0) = q(\x_1|\x_0) \prod_{t=2}^T q(\x_{t}|\x_{t-1}) = \prod_{t=1}^T q(\x_{t}|\x_{t-1}) \,.
\end{equation}
Using a Gaussian Markov diffusion kernel $K$, we have so-called Gaussian diffusion, as introduced in \cite{pmlr-v37-sohl-dickstein15}. Then, 
\Cref{eq:diffusion_kernel} becomes
\begin{equation}\label{eq:q-xt-from-xt-1}
    q({\x}_t | {\x}_{t-1}) = \N({\x}_t; \sqrt{1-\beta_t}\; {\x}_{t-1}, \beta_t \mathbf{I}) \,,
\end{equation}
where $\N$ denotes a Gaussian distribution and we have parameterized the distribution with mean $\bmu=\sqrt{1-\beta_t}\; {\x}_{t-1}$
and covariance matrix $\bSigma=\beta_t \mathbf{I}$.
The diffusion rate $\beta_t$ typically starts close to $0$ for small $t$, meaning that the variance of the conditional distribution in \Cref{eq:q-xt-from-xt-1} is small, while $\sqrt{1-\beta_t}$ is close to $1$, 
and thus the conditional mean is close to $\x_{t-1}$,  the sample at the previous step.
It also follows that 
\begin{equation}
\nonumber
    q(\x_T) \approx p_{\text{prior}}(\x_T) = \N(\x_T;\mathbf{0}, \mathbf{I}) \,,
\end{equation}
with $q(\x_T) = p_{\text{prior}}(\x_T)$ in the limit as $T\rightarrow\infty$.
A schematic visualisation of the diffusion process depicted as moving between distributions is shown in~\Cref{fig:diffusion-over-time}.

The Gaussian distribution-function in vector notation is
\begin{equation}
\nonumber
    \N(\x; \bmu, \bSigma) = \frac{1}{\sqrt{{\left(2\pi\right)}^k \det \bSigma}} 
    \text{exp}\left(-\frac{1}{2}\left(\x-\mathbf{\bmu}\right)^T\bSigma^{-1}\left(\x-\bmu\right)\right) \,.
\end{equation}
We can draw samples from this distribution using that if $\X\sim \N(\x; \bmu, \bSigma)$, then we can reformulate $\X$ as
\begin{equation}
\label{eq:multi_gauss_sample}
     \X = \bmu + \mathbf{AZ}, \quad \mathbf{Z} \sim \N(0,\mathbf{I}) \,,
\end{equation}
where $\mathbf{A}$ is choses to ensure that $\bSigma=\mathbf{AA}^T$.
Thus, defining $\alpha_t = 1-\beta_t$, we can use \Cref{eq:q-xt-from-xt-1} to write
\begin{equation}
\nonumber
    {\X}_t = \sqrt{\alpha_t} \, {\X}_{t-1}+\sqrt{1-\alpha_t} \,\Z_t \,,  \label{eq:t_from_t-1_and_std_gauss}
\end{equation}
if we let $\sqrt{1-\alpha_t}\,\mathbf{I}$ play the role of $\mathbf{A}$ in \Cref{eq:multi_gauss_sample} and choose  $\Z_t$ to follow the standard Gaussian distribution.
We can use this recursive relation to express ${\X}_t$ in terms of ${\X}_0$ directly as follows:
\begin{align}
    {\X}_t  &= \sqrt{\alpha_t} \,\X_{t-1}+\sqrt{1-\alpha_t} \, \Z_t \\
            &= \sqrt{\alpha_t \alpha_{t-1}}\,\X_{t-2} + \sqrt{\alpha_t(1-\alpha_{t-1})} \, \Z_{t-1} + \sqrt{1-\alpha_t} \Z_t \label{eq:fd_12} \\
            &= \sqrt{\alpha_t \alpha_{t-1}}\,\X_{t-2} + \sqrt{1-\alpha_t\alpha_{t-1}} \, \Z_{t-1:t} \label{eq:fd_2} \\
            &= \dots \label{eq:fd_3}   \\
            &= \sqrt{\Bar{\alpha}_t} \, \X_0+\sqrt{1-\Bar{\alpha}_t} \, \Z_{0:t} \, . \label{eq:fd_4} 
\end{align}
In \Cref{eq:fd_12} we have used \Cref{eq:t_from_t-1_and_std_gauss} to represent $\X_{t-1}$ via $\X_{t-2}$ and $\Z_{t-1}$. 
To get to \Cref{eq:fd_2} we used two facts about random variables: First, the sum of two Gaussian variables with parameters $(\bmu_1=\mathbf{0}, \bSigma_1=\sigma_1^2 \mathbf{I})$ and 
$(\bmu_2=\mathbf{0}, \bSigma_2=\sigma_2^2 \mathbf{I})$ is a new Gaussian variable with mean $\mathbf{0}$ and covariance $(\sigma_1^2+\sigma_1^2)\cdot \mathbf{I}$.
Second, for  a scalar $\gamma$ and a random variable $\mathbf{W}$, we have $\Var(\gamma\mathbf{W})=\gamma^2\,\Var(\mathbf{W})$.
Now, the transition from \Cref{eq:fd_12} to \Cref{eq:fd_2} holds because 
\begin{align}
    \sqrt{\alpha_t(1-\alpha_{t-1})} \Z_{t-1} + \sqrt{1-\alpha_t} \Z_t 
    &= \N(\mathbf{0}, \alpha_t(1-\alpha_{t-1})\mathbf{I}) + \N\mathbf{0},(1-\alpha_t)\mathbf{I})   \nonumber\\
    &= \N(\mathbf{0}, (1-\alpha_t \alpha_{t-1})\mathbf{I}) \nonumber\\
    &= \sqrt{1-\alpha_t \alpha_{t-1}} \N(\mathbf{0}, \mathbf{I}) \,. \nonumber
\end{align}

\begin{figure}
    \centering
    \includegraphics[width=0.45\textwidth]{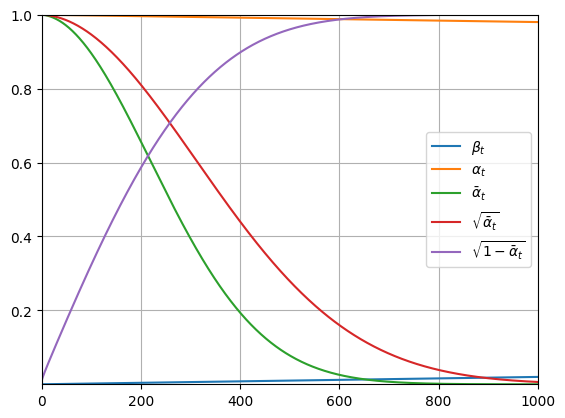}    \phantom{==}
    \includegraphics[width=0.45\textwidth]{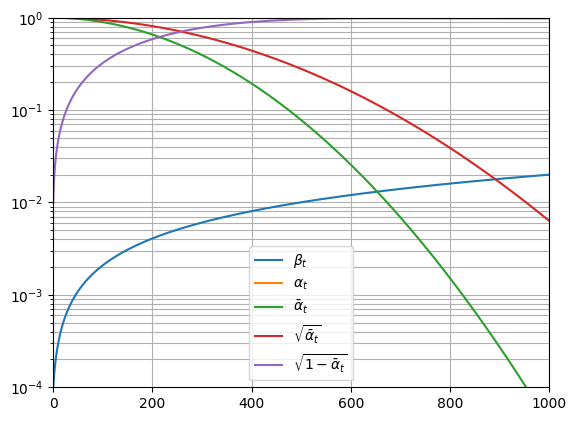}    
    \caption{\label{fig:diffusion_parameters} Parameter values for $\beta=[10^{-4},0.02]$ over $1000$ time steps $t$ using a linear schedule. 
    The information in the two figures  are the same, but the right-hand side uses log-scale on the $y$-axis to show the speed of which $\bar{\alpha}_t$ goes towards zero.}
\end{figure}

We have given the $\Z$-variables explicit indices in the development above to keep track of the time step(s), but remember that all of them follow the standard Gaussian distribution with mean $\mathbf{0}$ and covariance-matrix $\mathbf{I}$.
The dots in  (\ref{eq:fd_3}) indicate repeating this procedure for all steps until 
${\X}_0$, and in \Cref{eq:fd_4} we have introduced the compact notation $\Bar{\alpha}_t=\prod_{i=1}^t \alpha_i$. We have thus arrived at an expression for generating a single noisy sample ${\x}_t$ given an initial sample ${\x}_0$:
\begin{equation} \label{eq:q_t_0}
    q({\x}_t | {\x}_0) = \N\left({\x}_t; \sqrt{\Bar{\alpha}_t} {\x}_0, (1-\Bar{\alpha}_t) \mathbf{I}\right) \,.
\end{equation}
Note that the covariance matrix of the added noise is diagonal throughout.
The samples $\x_t$ gradually become more noisy, and as $T \to \infty$, ${\x}_T$ is drawn from an isotropic Gaussian distribution, $q({\x}_T|{\x}_0) \approx \N(0,\mathbf{I}) = p_{\text{prior}}$. \cite{ho_2020} use $T=1000$.
Regarding the variance schedule, values of $\beta_t$ are typically in the range $[10^{-4},0.02]$. 
The schedule used in \cite{pmlr-v37-sohl-dickstein15} defines a linear relationship between $t$ and $\beta_t$, while the one used in \cite{nichol_2021} is a cosine. The parameter values for this range, using a linear scale for simplicity, are shown in \Cref{fig:diffusion_parameters}.

To summarise, the forward diffusion process consists of a step-wise transformation of a sampled ${\x}_0$ to a random Gaussian noise-variable ${\x}_T \sim \N(0,\mathbf{I})$. 
We start from  $\x_0$, and at each time-step $t=1, \ldots, T$, make adjustments so that $\x_t$ has a mean that is gradually moved towards zero  and a variance that gradually increases towards unity as $t$ increases. 
This is most easily seen by studying the numerical values of the parameters in \Cref{eq:q_t_0} using \Cref{fig:diffusion_parameters}. 

Finally, 
we can also use \Cref{eq:q_t_0} to generate a sample $\x_t$  directly from an original sample $\x_0$ via
\begin{equation} \label{eq:xt-and-x0}
    \x_t = \sqrt{\Bar{\alpha}_t}\x_0 + \sqrt{1-\Bar{\alpha}_t} \,\beps \,,\;\; \beps\sim\N(\mathbf{0}, \mathbf{I}).
\end{equation}

\begin{figure}
    \centering
    \begin{tabular}{c}
        \includegraphics[width=.85\textwidth]{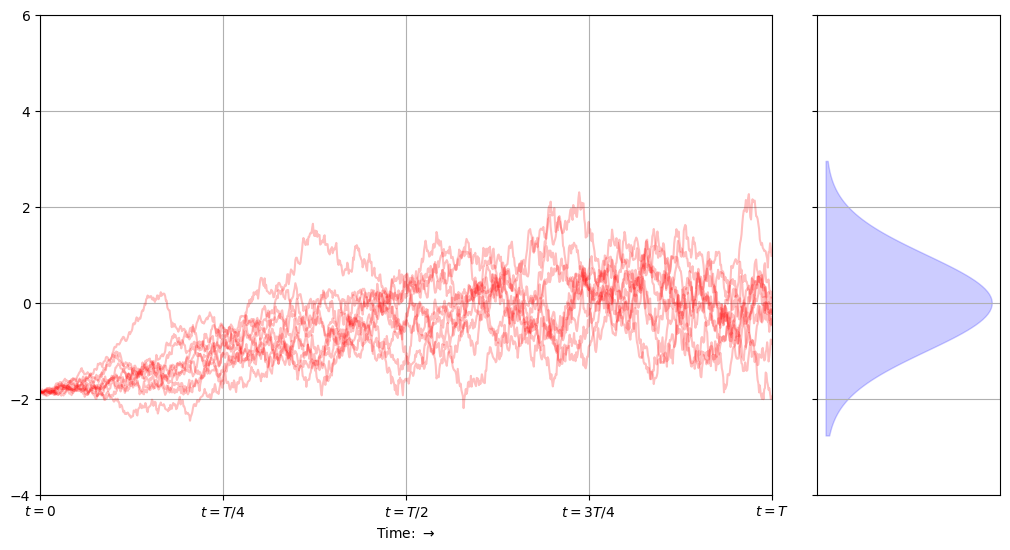}    \\
        \includegraphics[width=.85\textwidth]{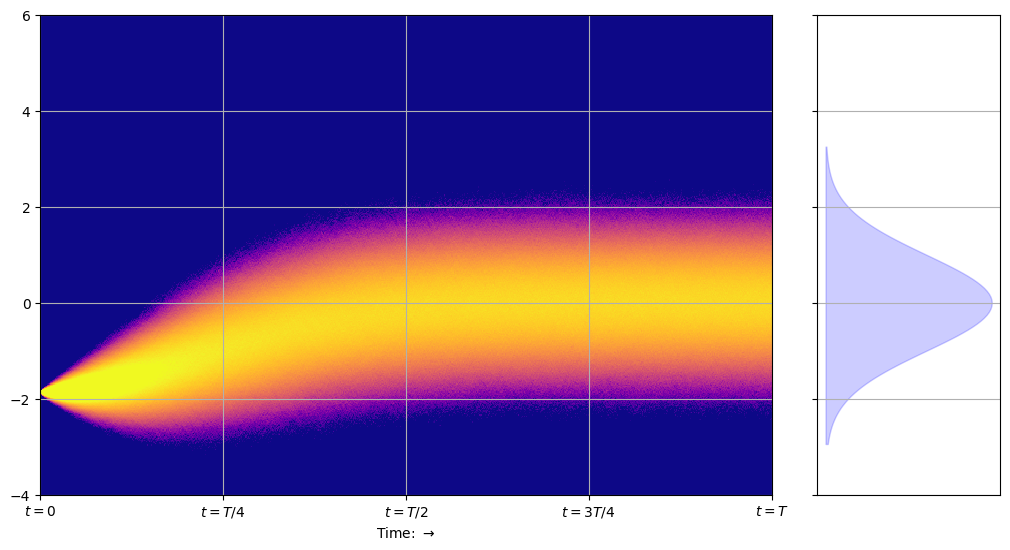}     
    \end{tabular}
    \caption{The forward process moves in the direction of increasing time (left to right). The marginal distribution on the right-hand side of each plot gives the distribution $q(\x_T|\x_0)$. 
    Top: $10$ trajectories sampled from the same starting-point $\x_0$. 
    Bottom:   For each $t$ the figure shows the estimated conditional distribution $q(\x_t|\x_0)$  based on $25.000$ trajectories sampled from the same starting-point $\x_0$.}
    \label{fig:forward_traces}
\end{figure}

\Cref{fig:forward_traces} shows example trajectories from a univariate diffusion process. 
The top part of the figure shows ten trajectories all starting from the same $\x_0$. The data is generated by iteratively sample using $q(\x_t|\x_{t-1})$ following \Cref{eq:q-xt-from-xt-1}.
Eventually, at $t=T$, the samples approximately follow the targeted Gaussian distribution shown on the right-hand side. 
The bottom part of \Cref{fig:forward_traces} gives the estimated density $q(\x_t|\x_0)$ for each $t$ based on a high number of sampled trajectories, see also \ \Cref{eq:q_t_0}, again using the same starting-point $\x_0$ as in the top part of the figure.

The last point to make about the forward diffusion process is that we do not need to train a machine learning model to do any of this. 
We are simply following \Cref{eq:q-xt-from-xt-1} mechanically to transform a sample $\mathbf{x}_0$  from the data distribution $p_\text{data}$ until we obtain a transformed sample $\mathbf{x}_T$  that (approximately) comes from the much simpler $p_\text{prior}$. 
One could ask what we have gained by this. That will hopefully become clear in the next subsection when we reverse the process: Starting from a sample $\mathbf{x}_T\sim p_{\text{prior}}$,
 we will gradually change that until we obtain an $\mathbf{x}_0$  that (approximately) is  from  $p_\text{complex}$.

\subsection{Generative modelling = undoing diffusion}
Having completed the forward diffusion process, we would now like to be able to undo that again, that is, gradually remove noise from a sample ${\x}_T$, ending up with a data sample $\x_0$ from the original distribution. 
An important result in this context is by \cite{feller_1949}, showing that in the limit  of infinitesimal step size, the true reverse process will have the same distributional form as the forward process.
In our case, we should thus expect the reversed process to be a series of random variables following the Gaussian distribution (since the forward-process only involved Gaussians).  
In essence it should thus be \textit{possible} to model the reverse process --  we only need to calculate the parameters of the Gaussians. 
Nevertheless, while the forward diffusion process is fixed, finding the reverse process requires quite some effort. 

Ideally, we would just calculate $q({\x}_{t-1} | {\x}_t)$ and sample step-wise from the distribution guiding us from a sample $\x_T\sim p_\text{prior}$ to a sample $\x_0\sim p_\text{complex}$.
However, using Bayes' rule  to calculate $q({\x}_{t-1} | {\x}_t)$ from $q({\x}_{t} | {\x}_{t-1})$ means that we will need to not only \textit{represent} the distribution $q(\x_0)$, but also be able to \textit{integrate} over it. 
This follows because the only way to use Bayes rule when we do not have access to the marginals $q(\x_{t-1})$ and $q(\x_{t})$
would be as follows:
\begin{eqnarray}
q(\x_{t-1}|\x_t) & = & \frac{q(\x_{t}|\x_{t-1})\cdot q(\x_{t-1})}{q(\x_{t})}\nonumber \\
& = & \frac{q(\x_{t}|\x_{t-1})\cdot \int_{\x_0} q(\x_{t-1}|\x_0)q(\x_0) d\x_0}{\int_{\x_0} q(\x_{t}|\x_0)q(\x_0) d\x_0}\, .\nonumber
\end{eqnarray}
Remember that we previously referred to $q(\x_0)$ as  $p_\text{complex}$ because we have no reason to believe that it is a Gaussian (or that it comes from any other mathematically convenient distributional family, for that matter): 
It is only for $t>0$ and conditionally on the starting-point $\x_0$ that all variables are Gaussians.  
In conclusion, we will not be able to calculate $q({\x}_{t-1} | {\x}_t)$ directly, but need to do something else. 
The idea is instead to find a suitable approximation to $q({\x}_{t-1} | {\x}_t)$ and let that approximation define the reverse process. 
Fortunately, we can define such a reverse diffusion process even if we limit our use of $p_\text{complex}$. In the following  we will only be accessing \textit{samples} from the distribution. 
We will generate those samples by randomly selecting examples from the training-data, which represents the data-distribution $p_{\text{data}}$, and therefore approximates $p_\text{complex}$.  
This is the core idea behind diffusion models, and how all of this fit together will be described next.

As an alternative to reversing the diffusion process with Bayes rule, we will define a new distribution that is meant to be representing the reverse process directly. 
That is, we will create a step-wise noise reduction process $p({\x}_{t-1}| {\x}_t)$ that at least approximates $q(\x_{t-1}|\x_t)$.
Leveraging the observation that the reverse process must have the same functional form as the forward process (given that each $\beta_t$ is sufficiently small in the forward process), each reverse step can be parameterised as a Gaussian,
and the parameters can be learned by fitting a neural network, as observed by \cite{pmlr-v37-sohl-dickstein15}.
This means that we only have to estimate the mean and variance of the distribution $p({\x}_{t-1}| {\x}_t)$ in order to draw samples from it.
Letting a neural network parameterised by $\btheta$ represent this distribution, i.e., produce the distributional parameters $\bmu$ and $\bSigma$,
we can thus write
\begin{equation}\label{eq:p_t-1_t}
    p_\btheta({\x}_{t-1}| {\x}_t) = \N\left({\x}_{t-1}; \bmu_{\btheta}({\x}_t, t), \bSigma_{\btheta}({\x}_t, t) \right) \,.
\end{equation}
When learning the network that generates the parameters in \Cref{eq:p_t-1_t}, the deep neural network takes as input the sample at time $t$, $\x_t$, in addition to the time step $t$ itself, in order to account for the variance schedule of the forward process; as different time steps are associated with different noise levels, the model must learn to undo these individually. 
%

Like the forward process, the reverse process is a Markov chain, and we can write the joint probability of a sequence of samples 
 as a product of conditionals and the marginal probability of ${\x}_T$,
\begin{equation}\label{eq:p_theta_joint}
    p_\btheta({\x}_{0:T}) 
    = p({\x}_T) \prod_{t=1}^T p_\btheta({\x}_{t-1}|{\x}_t) \,.
\end{equation}
Here, the distribution $p({\x}_T)$ is the same as $q({\x}_T)$, namely the pure noise distribution $p_{\text{prior}}=\N({\x}_T; 0,\mathbf{I})$. This  means that the generation process starts with Gaussian noise, followed by sampling from the learned individual steps of the reverse process.

\begin{figure}
    \centering
    \includegraphics[width=0.95\textwidth]{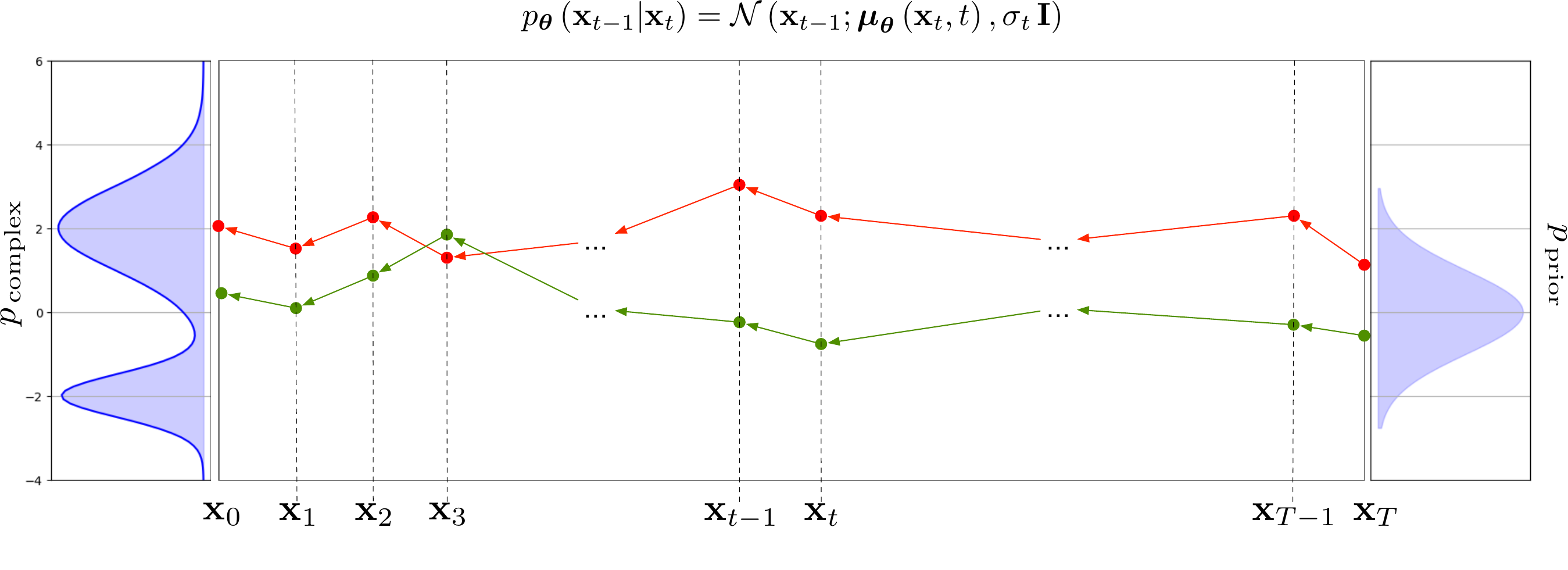}
    \caption{
    \label{fig:reverse-diffusion-over-time}
    The general idea of the reverse diffusion process is to go backwards in time, starting with a sample $\x_T\sim p_{\text{prior}}$ 
    and end up with $\x_0$ that (approximately) comes from $p_{\text{complex}}$.
    Compare this to the forward process in \Cref{fig:diffusion-over-time} to realise that the major difference between the forward 
    and the backward process is that when the 	forward process was  \textit{defined} by \Cref{eq:noising-process-q}, 
    the reverse needs to be \textit{learned}, cf. \Cref{eq:p_t-1_t}.
    }
 \end{figure}

Note that while the original data-samples (e.g., images) do not have zero off-diagonal covariance (neighbouring pixels contain information about each other), the noise added to the original samples was diagonal, meaning that we can assume that the variance of the removed noise is also diagonal, i.e.\ $\bSigma=\sigma\cdot\mathbf{I}$ for some scalar value $\sigma$. 
In the case of a diagonal covariance matrix $\bSigma$, the mean and the variance in each dimension can be estimated separately, and the multivariate density function can be described in terms of a product of univariate Gaussians. 
If the variance is given, we merely have to predict the mean. 
In \cite{nichol_2021}, it is shown empirically that learning the covariance can improve the quality of the generated samples, but this is not something we will focus on here.
We will rather predict only the mean of the reversed diffusion process-distribution, while the variance follows a schedule parameterised by $t$. 
The general process of the reverse process is thus as shown in \Cref{fig:reverse-diffusion-over-time}. 
We start by a sample from $p_\text{prior}$, use \Cref{eq:p_t-1_t} powered by the neural network with parameters $\btheta$, and eventually obtain $\x_0$. 
We will discuss two different reverse processes, namely the \textit{Denoising Diffusion Probabilistic Model} \cite{ho_2020,nichol_2021} and
\textit{Denoising Diffusion Implicit Models} \cite{ddim_arxiv}. However, we will first discuss how to define the loss function of the neural network. 

\subsection{The loss \label{sec:loss}}
What objective are we optimising when training the neural network to learn \Cref{eq:p_t-1_t}? 
All generative models attempt to learn the distribution of their training data, so it would make sense to maximise the likelihood assigned to ${\x}_0$ by the model.
Calculating this would require us to marginalise over all steps from $t=T$ down to $t=1$,
\begin{equation} \label{eq:marginal_p}
    p_\btheta({\x}_0) = \int p_\btheta({\x}_{0:T}) d{\x}_{1:T} \,.
\end{equation}
Maximising \Cref{eq:marginal_p} gives the process $p_\btheta$ over $\x_T\to\x_{T-1}\dots\x_1\to\x_0$ that has the largest log likelihood of producing the observed $\x_0$ from the noise $\x_T$. 
However, evaluating the above expression involves integrating over all possible trajectories from noise to the data manifold, which is intractable.
Instead we can maximise a lower bound of the log likelihood, taking a page out of the book of variational autoencoders.

To get to these results we will first discuss some results regarding variational inference (\Cref{sec:elbo}) and the VAE (\Cref{sec:VAE}). 
Unfortunately, some of the syntax used by the community behind these results differ from  what is used elsewhere in this document. 
Nevertheless, we have been true to the original lingo in our description, and then try to ``translate'' the core concepts and ideas back to our language in \Cref{sec:back_to_diff}.

\subsubsection{Variational Lower Bound}\label{sec:elbo}

\begin{figure}[ht]
    \centering
    \includegraphics[width=0.5\textwidth]{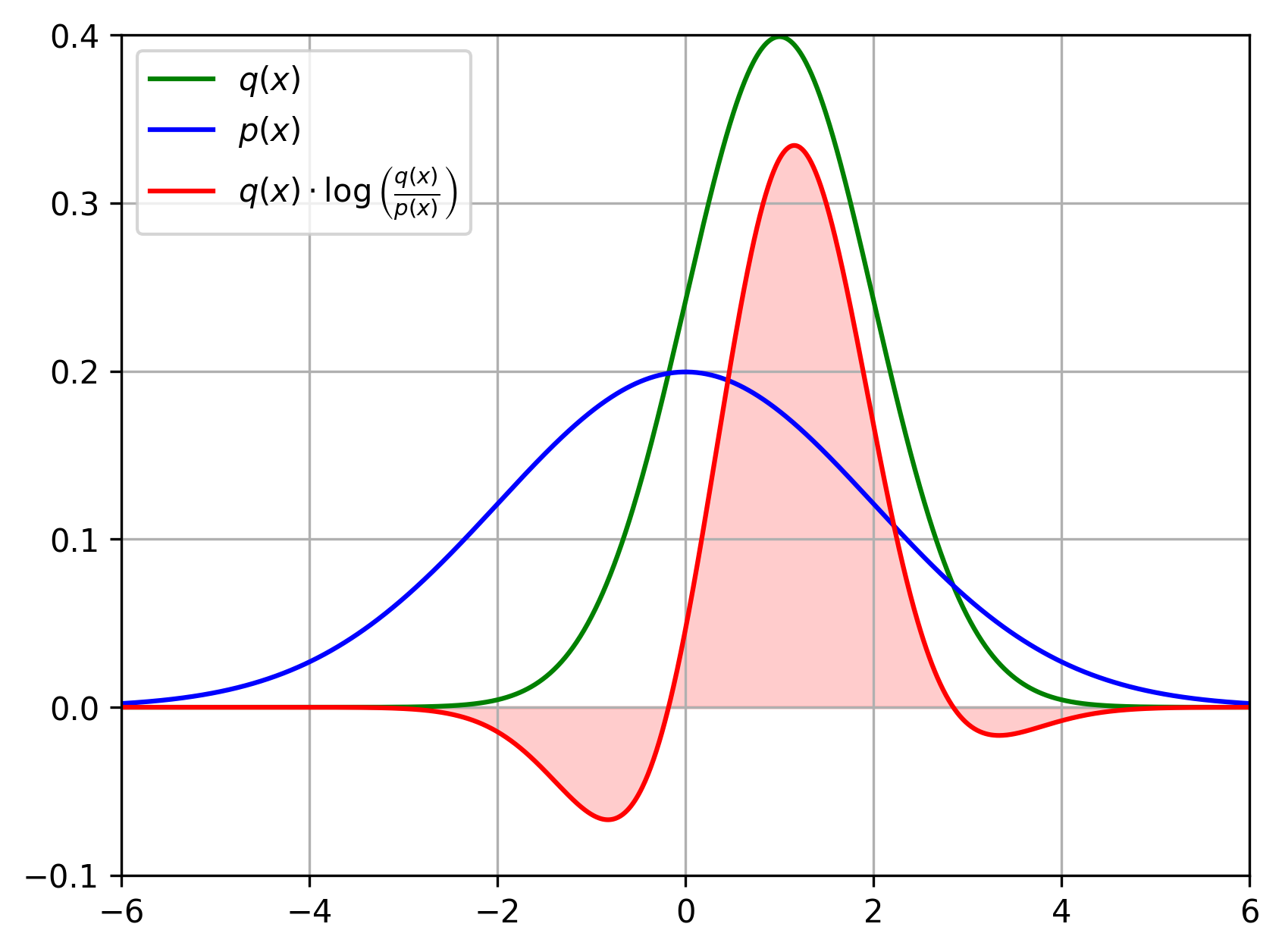}
    \caption{\label{fig:kl_div}The KL divergence between two Gaussians, here from  $q\sim\N(x; \mu=1,\sigma^2=1)$  (green) to $p\sim\N(x; \mu=0,\sigma^2=4)$ (blue) is given by the area under the red curve.}
\end{figure}

Imagine that we are given a process $\z \to \x$ that can generate data samples $\x$ from latent variables $\z$. The latent variable can e.g.\ contain information about the properties of an image, and through the process, denoted $p(\x|\z)$, the properties manifest into an actual image $\x$.
We would like to know the reverse process, i.e.\ how to obtain $\z$ from $\x$. Knowing the distribution of $\z$, we could try to use Bayes' rule,
\begin{equation}\nonumber
    p(\z | \x) = \frac{p(\z) p(\x|\z)}{\int p(\z,\x)d\z} \,.
\end{equation}
However, in general we do not know how to evaluate the integral $\int p(\z,\x)\,{\rm d}\z$. Therefore, we cannot calculate the denominator in the above expression, and instead choose to approximate $p(\z | \x)$ by a function $q_\bphi(\z|\x)$. This function can come from an approximation family $\mathcal{Q}$, where we seek to select the one that minimises some distance measure $\Delta$ between $q_\bphi$ and the true distribution $p$. Formally,
\begin{equation}\label{eq:optim_via_delta_min}
    \hat{q}_{\bphi}(\z|\x) = \arg \min_{q \in \mathcal{Q}} \Delta(q_\bphi(\z|\x) || p(\z|\x)) \,.
\end{equation}
So, if for instance we let $\mathcal{Q}$ be the set of all Gaussian distributions then $\phi=\{\bmu, \bSigma\}$ is the parameterisation that identifies each member of $\mathcal{Q}$, and 
\Cref{eq:optim_via_delta_min} abstractly describes how the best parameterisation is to be chosen. 
For reasons beyond our control, the Kullbach-Leibler (KL) divergence $\Dkl$ has been deemed a favourable distance measure, since it has some nice mathematical properties.
It is calculated as
\begin{equation}
\nonumber
    \DklP{q}{p} = \int q(\x) \log\frac{q(\x)}{p(\x)} \,\mathrm{d}\x= \mathbb{E}_{\x \sim q} \left[\log\frac{q(\x)}{p(\x)} \right] \,,
\end{equation}
and quantifies how different the probability distribution $p$ is from another distribution $q$. The shorthand $\mathbb{E}_{\x \sim q}$ indicates that the expectation is calculated using values of $\x$ coming from $q$. The KL divergence is always positive, but failing to be symmetrical under the interchange of $p$ and $q$ is one of its less desirable mathematical properties. \Cref{fig:kl_div} gives an example using two Gaussians, where the KL divergence is the area under the red curve. 
Note that the KL divergence cannot be negative, even if the red curve itself should sometimes  take negative values. 

Armed with the KL divergence, and using that the probability distribution of $q_\bphi(\z|\x)$ is normalised, i.e.\ $\int q_\bphi(\z|\x) d\z=1$,
 we can now rewrite the log likelihood  over the observed data, $\log p(\x)$, as
\begin{alignat*}{2}
    \log p(\x) 
    &= \log p(\x) \int q_\bphi(\z|\x) d\z &&  \text{Multiply by $\int q_\bphi(\z|\x) d\z=1$} \\
    &= \int q_\bphi(\z|\x) \log p(\x) d\z  && \text{Bring $p$ into integral}\\
    &= \E_{q_\bphi(\z|\x)} \left[\log p(\x) \right] && \text{Expectation} \\
    &= \E_{q_\bphi(\z|\x)} \left[\log \frac{p(\x,\z)}{p(\z|\x)} \right] && \text{Multiply by $\frac{p(\z|\x)}{p(\z|\x)}=1$; use chain rule} \\
    &= \E_{q_\bphi(\z|\x)} \left[\log \frac{p(\x,\z) q_\bphi(\z|\x)}{p(\z|\x) q_\bphi(\z|\x)} \right] && \text{Multiply by $\frac{q_\bphi(\z|\x)}{q_\bphi(\z|\x)}=1$} \\
    &= \E_{q_\bphi(\z|\x)} \left[\log \frac{p(\x,\z)}{q_\bphi(\z|\x)}\right]
    + \E_{q_\bphi(\z|\x)} \left[\log \frac{q_\bphi(\z|\x)}{p(\z|\x)} \right] \quad\quad && \text{Split up} \\
    &= \E_{q_\bphi(\z|\x)} \left[\log \frac{p(\x,\z)}{q_\bphi(\z|\x)}\right]
    + \DklP{q_\bphi(\z|\x)}{p(\z|\x)} && \text{Definition of KL divergence} \\
    &\geq \E_{q_\bphi(\z|\x)} \left[\log \frac{p(\x,\z)}{q_\bphi(\z|\x)}\right]
    && \text{KL divergence always positive} \,.
\end{alignat*}
The quantity on the last line is known as the Evidence Lower Bound (ELBO), and it is a \textit{lower bound} on $\log p(\x)$ because $\Dkl$ is always positive.
This step is often skipped in lecture notes and papers, referring to ``Jensen's inequality'', which generalises that the secant line of a convex function lies above the graph of the function itself.
In the standard formulation of variational autoencoders (VAEs), the objective is to maximise the ELBO. The term \textit{variational} refers to the optimisation of $q_\bphi(\z|\x)$ from the family $\mathcal{Q}$ of potential approximation functions.

Note that the true data distribution $p(\x)$ is constant with respect to $q_\bphi$ (the real world doesn't care which function we choose for approximating it), meaning that the ELBO and the KL divergence sum to a constant under optimisation of $\bphi$. Therefore, maximising the ELBO with respect to $\bphi$ is identical to a minimisation of $\Dkl$, corresponding to finding an optimal model for approximating the true latent distribution.
\subsubsection{Variational Autoencoders} \label{sec:VAE}
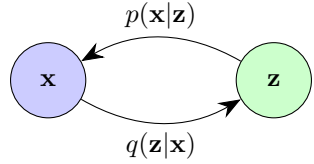
\begin{figure}
    \centering
    \begin{tikzpicture}
        \node[circle, draw, fill=blue!20, minimum size=10mm] (x) at (0, 0){$\x$};
        \node[circle, draw, fill=green!20, minimum size=10mm] (z) at (3, 0){$\z$};
        \draw[-{Stealth[length=3mm]}] (z) to [bend right=30] node [above, sloped] (TextNode1) {$p(\x|\z)$} (x);
        \draw [-{Stealth[length=3mm]}] (x) to [bend right=30] node [below, sloped] (TextNode1) {$q(\z|\x)$} (z);
    \end{tikzpicture}
    \caption{\label{fig:autoencoder}In a variational autoencoder, the encoder $q(\z|\x)$ defines a distribution over latent variables $\z$ given observations $\x$, and the decoder $p(\x|\z)$ defines the distribution over observations given the latent variables.}
\end{figure}
An autoencoder is a neural network consisting of an encoder $q$ that transforms observed data $\x$ to a latent representation $\z$, and a decoder $p$ that transforms the data from the latent representation back to the original, see \Cref{fig:autoencoder}.
We do not know the ground truth encoder $q(\z|\x)$ or the decoder $p(\x|\z)$, but we can estimate them using parameterised models $q_\bphi(\z|\x)$ and $p_\btheta(\x|\z)$, where the parameters $\bphi$ and $\btheta$ have to be optimised.

In order to optimise using the ELBO as loss function, we need to write it in terms we can calculate,
\begin{alignat*}{2}
    \log p(\x) 
    &\geq \E_{q_\bphi(\z|\x)} \left[\log \frac{p(\x,\z)}{q_\bphi(\z|\x)}\right] &&  \text{Definition of ELBO.}  \\
    &= \E_{q_\bphi(\z|\x)} \left[\log \frac{p_{\btheta}(\x|\z) p(\z)}{q_\bphi(\z|\x)}\right] && \text{Chain rule.} \\
    &= \E_{q_\bphi(\z|\x)} \left[\log p_{\btheta}(\x|\z)\right] + \E_{q_\bphi(\z|\x)} \left[\log \frac{p(\z)}{q_\bphi(\z|\x)}\right] \quad\quad && \text{Split up.} \\
    &= \underbrace{\E_{q_\bphi(\z|\x)} \left[\log p_{\btheta}(\x|\z)\right]}_{\text{reconstruction term}}
    - \underbrace{\DklP{q_\bphi(\z|\x)}{p(\z)}}_{\text{prior matching term}} && \text{Definition of KL divergence} \,.
\end{alignat*}
When optimising the ELBO we thus learn two distributions; the prior matching term optimises the parameters $\bphi$ of the encoding function $q_\bphi(\z|\x)$ to match the true latent distribution $p(\z)$, i.e.\ it encourages the approximate posterior to be similar to the prior on the latent variable, while the reconstruction term optimises the parameters $\btheta$ of the decoding function $p_\btheta(\x|\z)$, maximising the expected probability density assigned to the data given the latent variables. Optimising the parameters $\bphi$ and $\btheta$ jointly is a defining feature of VAEs.

In order to calculate the loss for each choice of parameters, we have to specify a prior on the latent variables, and for VAEs (as for Gaussian diffusion), the common choice is a standard multivariate Gaussian,
\begin{equation}\nonumber
    p(\z) = \N(\z; \mathbf{0}, \mathbf{I}) \,.
\end{equation}
During training, we get data samples $\x$ by drawing from our training data. Modelling the encoder as a multivariate Gaussian with unknown mean $\bmu_\bphi$  and covariance $\sigma_\bphi$, assuming diagonal covariance,
\begin{equation}
\nonumber
    q_\bphi(\z|\x) = \N(\z, \bmu_\bphi(\x), \sigma^2_\bphi(\x)\mathbf{I}) \,,
\end{equation}
the KL divergence term can be computed analytically. The reconstruction term can be approximated using a Monte Carlo (MC) estimate (which is a fancy way of saying ``random sampling''), see \Cref{sec:mc_estimator}. 
We thus rewrite our objective as 
\begin{equation}
\nonumber
        \arg\max_{\bphi,\btheta} \E_{q_\bphi(\z|\x)} \left[\log p_{\btheta}(\x|\z)\right] - \DklP{q_\bphi(\z|\x)}{p(\z)}
        \approx
        \arg\max_{\bphi,\btheta} \sum_{l=1}^L \log p_{\btheta}(\x|\z^l) - \DklP{q_\bphi(\z|\x)}{p(\z)} \,,
\end{equation}
where the Monte Carlo action happens as we sample the $\z^l$ from $q_\bphi(\z|\x)$ for every $\x$ in the training data.
After training, new data can be generated by sampling directly from the latent space via $p(\z)$ and running the decoder on these samples. 

\subsubsection{Back to the diffusion}\label{sec:back_to_diff}
The diffusion model story is the reverse of the variational autoencoder story from \Cref{sec:VAE}: We start with an object $\x_0$ that we gradually convert to noise through the known process $q(\x_t|\x_{t-1})$. This is in contrast to the VAEs setting, as our forward process -- corresponding to the encoding to latent variables -- is not learned, but fixed. We therefore do not have a parameterised $q_\bphi$.
However, the reverse process has to be learned, so we still have the parameterised $p_\btheta$. 
We thus only need one neural network model (that learns the reverse process), analogous to the decoder part of the VAE. 
Nevertheless, we can make use of the training objective used by VAEs.

We have the conditional distribution from \Cref{eq:noising-process-q}, repeated here for convenience:
\begin{equation} \label{eq:q_process_repeated}
    q(\x_{1:T}|\x_0) = \prod_{t=1}^T q(\x_{t}|\x_{t-1}) \,,
\end{equation}
and the joint distribution from \Cref{eq:p_theta_joint},
\begin{equation}\label{eq:p_process_repeated}
    p(\x_{0:T}) = p(\x_T) \prod_{t=1}^{T} p_\btheta(\x_{t-1} | \x_t) \,,
\end{equation}
where, as before, $p(\x_T)=\N(\mathbf{0},\mathbf{I})$.
Let us now derive the ELBO again, using the diffusion model notation and these two distributions:
\begin{alignat}{2}
    \log p(\x_0) &= \log \int p(\x_{0:T}) d\x_{1:T} && \nonumber \\
    &= \log \int \frac{p(\x_{0:T})q(\x_{1:T}|\x_0)}{q(\x_{1:T}|\x_0)} d\x_{1:T} && \nonumber \\
    &= \log \E_{q(\x_{1:T}|\x_0)} \left[ \frac{p(\x_{0:T})}{q(\x_{1:T} | \x_0)} \right] &&  \nonumber \\
    & \geq \E_{q(\x_{1:T}|\x_0)} \left[ \log \frac{p(\x_{0:T})}{q(\x_{1:T} | \x_0)} \right]
    && \text{Jensen's inequality} \nonumber \\
    &= \E_{q(\x_{1:T}|\x_0)} \left[ \log \frac{p(\x_T) \prod_{t=1}^{T} p_\btheta(\x_{t-1}|\x_t)}{\prod_{t=1}^{T}q(\x_t | \x_{t-1})} \right] \phantom{==}
    && \text{Use \Cref{eq:q_process_repeated} and \Cref{eq:p_process_repeated}.} 
    \label{eq:diffusion_elbo}
\end{alignat}
We will later promote $\log p(\x_0)$ to be the (negative) loss function, so let us begin by discussing why this is reasonable. Obviously, $\log p(\x_0)$ evaluated at an element from the training data set gives the log likelihood of that specific example, i.e., the log likelihood that the reverse process starting from a random initialisation should end up at that specific data point. If this number is high, we have a reverse process that makes generating the data point likely. In other words, the reverse process has a good chance of producing $\x_0$.
If the reverse process is able to give a reasonably sized log likelihood to all the examples in our training data, this indicates that the reverse process is well suited to generate our data. 
Overall, using the negative log likelihood summed over all data points in the training data set seems reasonable, and is therefore the strategy we would like to follow. 
Unfortunately, this quantity is not available to us, which is why we instead use the ELBO discussed in \Cref{sec:elbo} and defined for diffusion models in \Cref{eq:diffusion_elbo}. 

    Next, we look at how the ELBO can be calculated efficiently. In the following, we use that for positive real numbers $a$ and $b$, $\log a\cdot b = \log a+\log b$, and $\log a/ b = \log a-\log b$. Using this in \Cref{eq:diffusion_elbo}, we obtain
\begin{align}
    \log p_\btheta(\x_0) 
    & \geq 
    \E_{q(\x_{1:T}|\x_0)} \left[ \log \frac{p(\x_T) \prod_{t=1}^{T} p_\btheta(\x_{t-1}|\x_t)}{\prod_{t=1}^{T}q(\x_t | \x_{t-1})} \right]  
    \nonumber \\
    & = 
    \E_{q(\x_{1:T}|\x_0)} \left[\log p_\btheta(\x_0|\x_1)\right] + 
    \E_{q(\x_{1:T}|\x_0)} \left[\log \frac{p_\btheta(\x_T)}{q(\x_T|\x_{T-1})}\right] + 
    \sum_{t=1}^{T-1}
    \E_{q(\x_{1:T}|\x_0)} \left[\log \frac{p_\btheta(\x_t|\x_{t+1})}{q(\x_t|\x_{t-1})}\right]
    \nonumber \\
    & = 
    \E_{q(\x_{1}|\x_0)} \left[\log p_\btheta(\x_0|\x_1)\right] + 
    \E_{q(\x_{T-1:T}|\x_0)} \left[\log \frac{p_\btheta(\x_T)}{q(\x_T|\x_{T-1})}\right] 
    \nonumber \\
    &  \phantom{=======}
    + \sum_{t=1}^{T-1}
    \E_{q(\x_{t-1:t+1}|\x_0)} \left[\log \frac{p_\btheta(\x_t|\x_{t+1})}{q(\x_t|\x_{t-1})}\right].
    \label{eq:soon_fav}
\end{align}
The last equality follows from that the expectation of a function of the subset of variables $\x_{a:b}$ with $1\leq a \leq b \leq T$, say $f(\x_{a:b})$ wrt.\ a distribution $q(\x_{1:T}|\x_0)$, is given by taking the expectation only over the variables $\x_{a:b}$: 
\begin{align}
\E_{q(\x_{1:T}|\x_0)} \left[f(\X_{a:b})\right] 
&= 
\int_{\x_{1:T}} q(\x_{1:T}|\x_0)  \cdot f(\x_{a:b}) \, d\x_{1:T} \nonumber\\
&=
\int_{\x_{1:T}} q(\x_{a:b}|\x_0) \cdot q(\x_{1:a-1, b+1:T}|\x_0, \x_{a:b})  \cdot f(\x_{a:b}) \, d\x_{1:T} \nonumber\\
&=
\int_{\x_{a:b}}  q(\x_{a:b}|\x_0) f(\x_{a:b}) 
\underbrace{\int_{\x_{1:a-1, b+1:T}} \hspace*{-35pt} q(\x_{1:a-1, b+1:T}|\x_0, \x_{a:b})
\, d\x_{1:a-1, b+1:T}}_{\text{The inner integral equals $1$.}}
\,d\x_{a:b} \nonumber\\
&=
\int_{\x_{a:b}}  q(\x_{a:b}|\x_0) f(\x_{a:b}) 
\,d\x_{a:b} \nonumber\\
&= 
\E_{q(\x_{a:b}|\x_0)} \left[f(\X_{a:b})\right]. \label{eq:simplified-expectations-for-kl}
\end{align}

With the help of \Cref{eq:simplified-expectations-for-kl}, and remembering that $\DklP{q}{p_\btheta}=\E_{\x\sim q}\left[\log\frac{q(\x)}{p_\btheta(\x)}\right]$, 
we can simplify \Cref{eq:soon_fav}
as follows:
\begin{align}
    \log p_\btheta(\x_0) 
     & \geq 
    \overbrace{
        \E_{q(\x_{1}|\x_0)} \left[\log p_\btheta(\x_0|\x_1)\right] 
    }^{\text{Reconstruction}}
    -
    \overbrace{
        \E_{q(\x_{T-1}|\x_0)} \left[\DklP{q(\x_T|\x_{T-1})}{p_\btheta(\x_T)}\right]
    }^{\text{Prior matching}}
    \nonumber\\ & \phantom{=========}
    - \sum_{t=1}^T
    \underbrace{
        \E_{q(\x_{t-1,t+1}|\x_0)} \left[
            \DklP{q(\x_t|\x_{t-1})}{p_\btheta(\x_t|\x_{t+1})}
        \right]
    }_{\text{Consistency terms}} \,,
    \label{eq:fav}
\end{align}
which is finally the (negative) loss function for training our diffusion model.

\begin{figure}
     \centering     
     \begin{subfigure}[b]{0.45\textwidth}
         \centering
         \includegraphics[width=\textwidth]{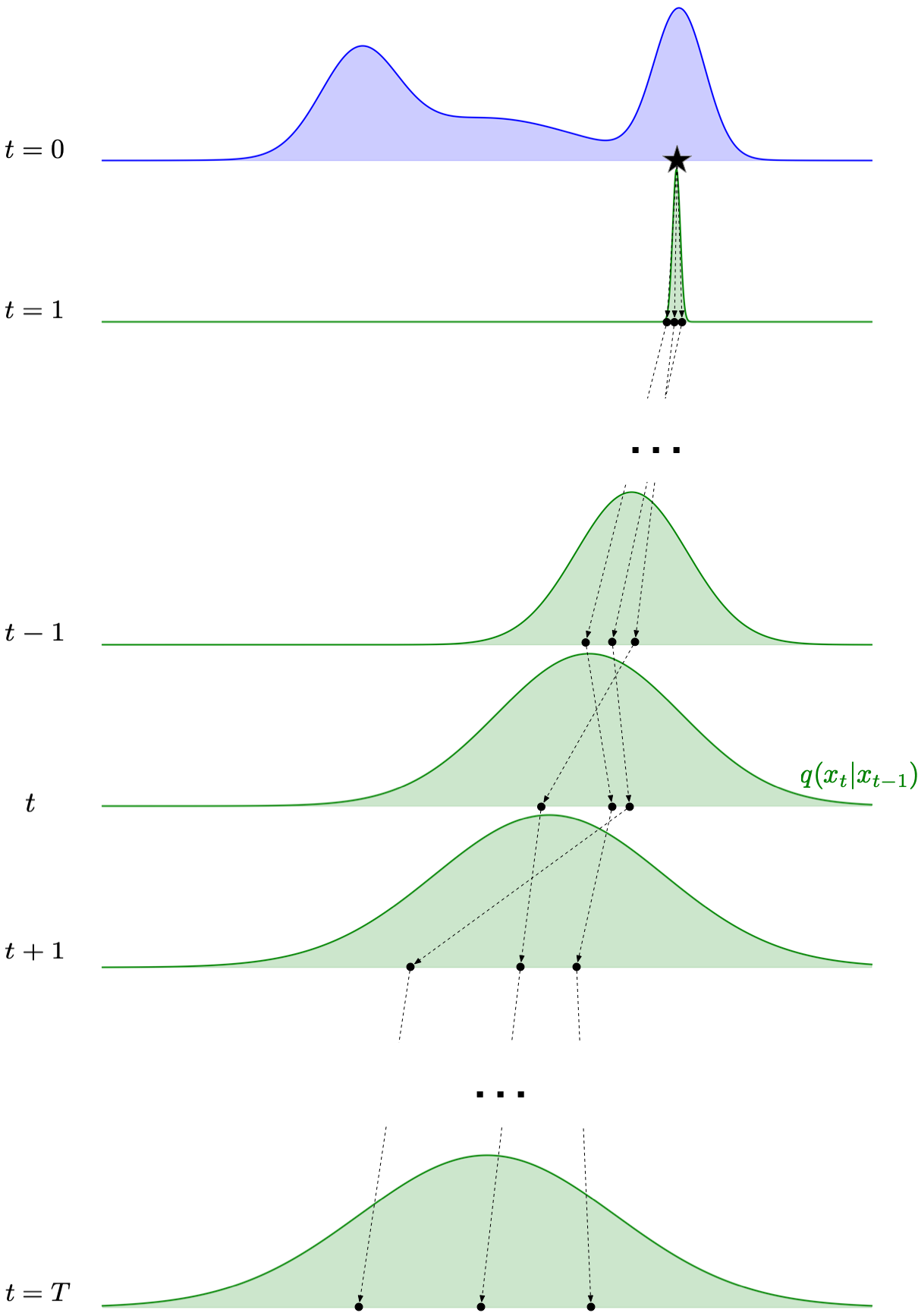}    
         \caption{Forward process: $q(\x_{1:T}|\x_{0})$}
         \label{fig:fav_forward_proc}
     \end{subfigure}
     \hfill
     \begin{subfigure}[b]{0.45\textwidth}
         \centering
         \includegraphics[width=\textwidth]{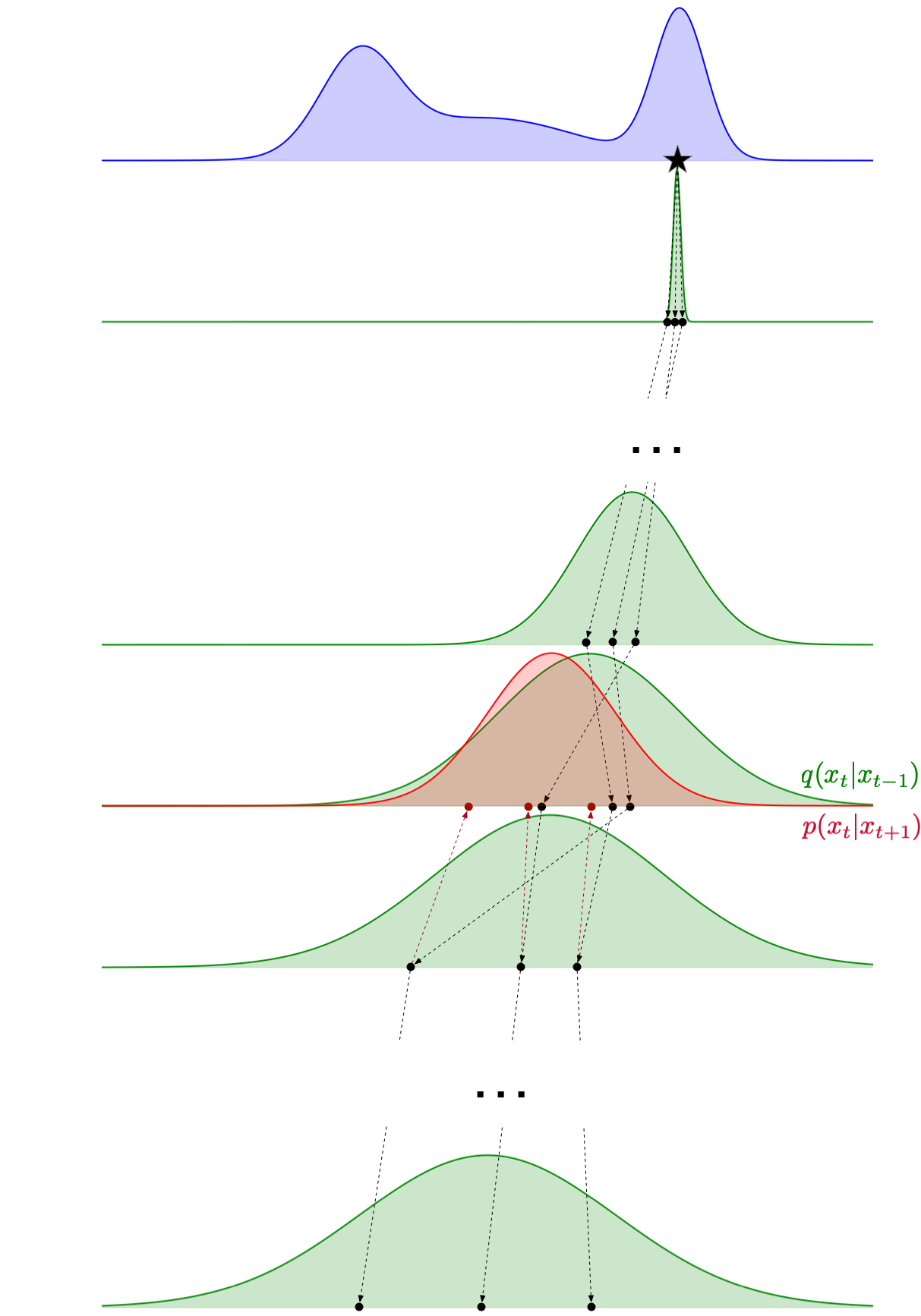}    
         \caption{Consistency term  in \Cref{eq:fav}}
         \label{fig:fav_final}
     \end{subfigure}
     \caption{Motivation for the consistency term in \Cref{eq:fav}.}
     \label{fig:fav}
\end{figure}

Let us look at each term of \Cref{eq:fav} in turn: The reconstruction term represents how successful the model is on average at reconstructing a data point $\x_0$ after one step in the diffusion process. 
The prior matching term compares the average result of the forward process at $\x_T$ to $p_\btheta(\x_T)$. 
Since 
$p_\btheta(\x_T)$ is fixed to be the standard Gaussian already, the prior matching term will be disregarded in the following. 
This leaves us with the consistency term, which we will consider in detail with the help of \Cref{fig:fav}.

Starting with \Cref{fig:fav_forward_proc}, this diagram shows the (forward) diffusion process evaluated from an initial $\x_0$. We sample from $q(\x_t|\x_{t-1})$ at each time-step $t$, and end up at $\x_T$. 
The figure indicates only a handful of trajectories using 
$q(\x_t|\x_{t-1})\sim\mathcal{N}\left(\x_t; \sqrt{1-\beta_t}\, \x_{t-1}, \beta_t\,\mathbf{I}\right)$ (see \Cref{eq:q-xt-from-xt-1}), 
but we can of course perform this as many times as we want. As defined in \Cref{eq:q_t_0}, we also know that 
$q(\x_t|\x_0)\sim\mathcal{N}\left(\x_t; \sqrt{\Bar{\alpha}_t}\, \x_0, \left(1-\Bar{\alpha}_t\right)\mathbf{I}\right)$. 

In \Cref{fig:fav_final}, we focus on a given $t$ and can reason as follows: 
\begin{quote}
\textit{``If the reverse process $p(\x_{t}|\x_{t+1})$ works well at time $t$,  we should be able to sample from $p(\x_{t}|\x_{t+1})$ and get  values of $\x_t$ similar to what we got when sampling using the forward process $q(\x_t|\x_{t-1})$.''}    
\end{quote}
This means that the distribution for the reverse process at time $t$  (red area in \Cref{fig:fav_final}) should be as close as possible to the (green) forward-process at the same time. 
 
As indicated in \Cref{fig:fav_final}, one idea to operationalise this insight is to take the samples obtained by the forward process from $\x_0$ at time $t+1$, and use those to start off the sampling of  $p(\x_{t}|\x_{t+1})$. The resulting distribution should be compared to the forward process that we can obtain by looking at samples from$q(\x_t|\x_{t-1})$ initialized at time $t-1$ with samples obtained by starting from $\x_0$. 

Armed with this intuition, we turn to defining $p(\x_{t}|\x_{t+1})$. First, and as already noted, we know that the \textit{forward} process will be a Gaussian distribution.
Since the \textit{reverse} is supposed to mimic that distribution over $\x_t$, it is reasonable to enforce that the reverse process should be Gaussian, too. 
Finally, we will use the KL divergence as a means to compare distributions, and we now aim to formalise the general idea. 
To minimise $\DklP{q(\x_t|\X_{t-1})}{p_\btheta(\x_t|\X_{t+1})}$ we must let the two conditioning variables $\X_{t-1}$ and $\X_{t+1}$ be related to our initial $\x_0$. 
In principle we can think that we are sampling the forward process starting from $\x_0$ a number of times and average  $\DklP{q(\x_t|\X_{t-1})}{p_\btheta(\x_t|\X_{t+1})}$ over these samples. Mathematically we here use that $\X_{t-1}\sim q(\x_{t-1}|\x_0)$ and $\X_{t+1}\sim q(\x_{t+1}|\x_0)$, and we end up with  $\E_{q(\X_{t-1,t+1}|\x_0)} \left[\DklP{q(\X_t|\X_{t-1})}{p_\btheta(\X_t|\X_{t+1})}\right]$ as our measure of the quality of $p_\btheta$ at time $t$. 
This is exactly the consistency term at time $t$ in \Cref{eq:fav}.
Notice that since both $p_\btheta$ and $q$ are Gaussian distributions, we can calculate the KL divergence analytically (see \Cref{eq:KLnormal} below). 

\subsubsection{Only one expectation}
As briefly eluded to above, it is natural to define our loss function as $\Loss(\btheta)=-\log p_\btheta(\x_0)$ because a low log likelihood of an observed data-point $\x_0$, $\log p_\btheta(\x_0)$, implies poor predictive performance for that $\x_0$. 
Since we are unable to calculate this log likelihood analytically, we can instead utilise \Cref{eq:fav}, repeated here for ease of reference:
\begin{align}
    \log p_\btheta(\x_0) 
      &\geq 
        \E_{q(\x_{1}|\x_0)} \left[\log p_\btheta(\x_0|\x_1)\right] 
    -  \E_{q(\x_{T-1}|\x_0)} \left[\DklP{q(\x_T|\x_{T-1})}{p_\btheta(\x_T)}\right] \nonumber \\ &
    - \sum_{t=1}^T
        \E_{q(\x_{t-1,t+1}|\x_0)} \left[
            \DklP{q(\x_t|\x_{t-1})}{p_\btheta(\x_t|\x_{t+1})}
        \right]
    \nonumber
\end{align}

Notice here the term 
$ 
\E_{q(\x_{t-1,t+1}|\x_0)} \left[
            \DklP{q(\x_t|\x_{t-1})}{p_\btheta(\x_t|\x_{t+1})}
        \right]
$, 
where we for each $t$ need to calculate the expectation over $\X_{t-1,t+1}|\x_0$. The source for the double expectation is that while the $q$-process works in the forward direction (from $t-1$ to $t$), the $p_\btheta$-process works in reverse (from $t+1$ to $t$). 
The idea now is to use Bayes rule to change the direction of the $q$-process in our definition of the loss function, meaning that we try to express  
\begin{equation}
q(\x_{1:T}|\x_0) = \prod_{t=1}^Tq(\x_{t}|\x_{t-1})
\nonumber
\end{equation}
through the alternative representation where we now reverse the process, i.e., focus on the transition from $t$ to $t-1$ instead of the forward process: 
\begin{equation}
q(\x_{1:T}|\x_0) = q(\x_T|\x_0) \prod_{t=2}^{T} q(\x_{t-1}|\x_{t}, \x_0)\,.
\nonumber
\end{equation}
Ensuring that both processes  $p_\btheta$ and $q$ are reversed in time makes them easier to compare.
Doing so, we find that the loss can be expressed as $-\log p_\btheta(\x_0) \leq \mathcal{L}_\text{vlb} \coloneqq \sum_{t=0}^T \Loss_t$ with 
\begin{align} 
    \Loss_0 &= -\E_{q(\x_{1}|\x_0)} \left[\log p_\btheta(\x_0|\x_1) \right] \nonumber\\
    \Loss_{t-1} &= \E_{q(\x_{t}|\x_0)} \left[\DklP{q(\x_{t-1}|\x_t,\x_0)}{p_\btheta(\x_{t-1}|\x_t)}\right],~ \text{ for } 2\leq t\leq T 
    \label{eq:loss_t-1}\\
    \Loss_T  &= \DklP{q(\x_T|\x_0) }{ p(\x_T)} \nonumber \,,
\end{align}
where you should notice that the $p(\x_T)$ used in $\Loss_T$ is parameter-free (no $\btheta$) as it is simply assumed to be a standard Gaussian.

Calculating $\Loss_0$ and $\Loss_T$ is straightforward, and -- as above -- we argue that $\Loss_T$ can be neglected.
To evaluate \Cref{eq:loss_t-1}, we calculate $q(\x_{t-1}|\x_t,\x_0)$ using Bayes' rule:
\begin{equation}
\label{eq:bayes-rule-for-qt-1}
    q(\x_{t-1}|\x_t,\x_0) = \frac{q(\x_{t-1}|\x_0) q(\x_{t}|\x_{t-1},\x_0)}{q(\x_t|\x_0)} \, .
\end{equation}
We already know how to calculate two of the terms in \Cref{eq:bayes-rule-for-qt-1} (see \Cref{eq:q_t_0}), which are 
\begin{align}
    q(\x_t|\x_0) &= \N(\x_t;\sqrt{\Bar{\alpha}_{t}} \x_0, (1-\Bar{\alpha}_t)\mathbf{I})\, , \nonumber\\
    q(\x_{t-1}|\x_0) &= \N(\x_{t-1};\sqrt{\Bar{\alpha}_{t-1}} \x_0, (1-\Bar{\alpha}_{t-1})\mathbf{I})\, . \nonumber
\end{align}
For the final term, we utilise that $q(\x_{1:T}|\x_0)$ is Markovian so that $q(\x_t|\x_{t-1},\x_0) = q(\x_t|\x_{t-1})$, defined in \Cref {eq:q-xt-from-xt-1}, so that
\begin{equation}
\nonumber
    q(\x_t|\x_{t-1},\x_0) = q(\x_t|\x_{t-1})= \N({\x}_t; \sqrt{1-\beta_t} {\x}_{t-1}, \beta_t \mathbf{I}) \label{eq:q-xt-from-xt-1-and-x0}.
\end{equation}

Now, we have defined all the terms used to calculate $q(\x_{t-1}|\x_t,\x_0)$ in \Cref{eq:bayes-rule-for-qt-1}, and some pencil pushing reveals that
\begin{equation}    \label{eq:q-xt-from-xt-1-and-x0-with-tildes}
    q(\x_{t-1}|\x_t,\x_0)  = \N\left(\x_{t-1}; \; \Tilde{\mu}(\x_t,\x_0), \;\Tilde{\beta}_t \mathbf{I}\right) , 
\end{equation}
with 
\begin{equation}\label{eq:q_reverse_process}
    \Tilde{\mu}_t(\x_t, \x_0) \coloneqq \frac{\sqrt{\Bar{\alpha}_{t-1}} \beta_t}{1-\Bar{\alpha}_t} \x_0 + \frac{\sqrt{\alpha_t}(1-\Bar{\alpha}_{t-1})}{1-\Bar{\alpha}_t} \x_t 
     \,;\quad
    \Tilde{\beta}_t \coloneqq \frac{1-\Bar{\alpha}_{t-1}}{1-\Bar{\alpha}_t} \beta_t
    .
\end{equation}

To summarise, the loss can be decomposed as $\Loss = \sum_{t=0}^T \Loss_t$, and if we zoom in on the contributions $2\leq t\leq T$, then $\Loss_{t-1}$  is determined by a KL divergence term between two Gaussian distributions: $q(\x_{t-1}|\x_t,\x_0)$, that we now know how to calculate, and $p_\btheta(\x_{t-1}|\x_t)$,  that we will choose ourselves, i.e., optimise with respect to the parameters $\btheta$. 

Moving on, we note that the KL divergence between two multivariate Gaussian distributions, $\DklP{\N(\bmu_q,\bSigma_q)}{\N(\bmu_p,\bSigma_p)}$, can be calculated as
\begin{equation}  
    \DklP{\N(\bmu_q,\bSigma_q)}{\N(\bmu_p,\bSigma_p)} = \frac{1}{2}\left[
    \log\frac{|\bSigma_p|}{|\bSigma_q|} - d + \text{tr}(\bSigma_p^{-1}\bSigma_q)
    + \left(\bmu_q-\bmu_p\right)^\top \bSigma_p^{-1} \left(\bmu_q-\bmu_p\right)
    \right]\,.
    \label{eq:KLnormal}
\end{equation}
We want to minimise this KL divergence, and can choose both $\bmu_p$ and $\bSigma_p$ freely when doing so. 
First, note that we know $\bSigma_q=\tilde{\beta}_t\textbf{I}$, and choose $\bSigma_p$ to be the same. 
This gives us 
\begin{align}
    \DklP{\N(\bmu_q,\bSigma_q)}{\N(\bmu_p,\bSigma_p)} & = \frac{1}{2}\left[\left(\bmu_q-\bmu_p\right)^\top \bSigma_q^{-1} \left(\bmu_q-\bmu_p\right) \right] \nonumber \\
    &=
    \frac{1}{2\tilde{\beta}_t}\left|\left|\bmu_q-\bmu_p\right|\right|_2^2 \, .
    \label{eq:KLnormal-simpler}
\end{align}  

When minimising $\Loss_{t-1}$, we would therefore prefer to choose $p_\btheta(\x_{t-1}|\x_t)$ as a Gaussian with mean $\Tilde{\bmu}\left(\x_t,\x_0\right)$ 
to perfectly match $q(\x_{t-1}|\x_t,\x_0)$ in \Cref{eq:q-xt-from-xt-1-and-x0-with-tildes}. 
However,
we are not able to calculate $\Tilde{\bmu}\left(\x_t,\x_0\right)$,  since the $p$-distribution is only conditioned on $\x_t$, while $\x_0$ is unknown. Basically, what we  try to do is force the $p_\btheta$ - distribution that \textit{is not} conditioned on $\x_0$ to fit with the $q$-distribution that \textit{is} conditioned on $\x_0$ as we minimise $\Loss_{t-1}$.  

The solution is to train a neural network to guess what $\x_0$ is. Basically, we create a neural network with parameters $\btheta$ that 
takes as input the noisy object $\x_t$ and the time-index $t$, and outputs a best guess on the final $\x_0$. We use $\hat\x_\btheta(\x_t, t)$ 
to denote this ``predicted $\x_0$''. 

Using 
\begin{alignat}{2}
    \nonumber  
    \bmu_q &= \frac{\sqrt{\Bar{\alpha}_{t-1}} \beta_t}{1-\Bar{\alpha}_t} \x_0 + \frac{\sqrt{\alpha_t}(1-\Bar{\alpha}_{t-1})}{1-\Bar{\alpha}_t} \x_t  && \text{(From \Cref{eq:q_reverse_process})}\\  
    \nonumber
    \bmu_p &= \frac{\sqrt{\Bar{\alpha}_{t-1}} \beta_t}{1-\Bar{\alpha}_t} \hat\x_\btheta(\x_t, t) + \frac{\sqrt{\alpha_t}(1-\Bar{\alpha}_{t-1})}{1-\Bar{\alpha}_t} \x_t \phantom{==}  && \text{(Substituting $\hat\x_\btheta(\x_t, t)$ for the unknown $\x_0$)}
\end{alignat}
in \Cref{eq:KLnormal-simpler}, yields
\begin{equation} \label{eq:loss_with_weighs}
    \Loss_{t-1} = \frac{1}{2\Tilde{\beta}_t} \; \frac{\bar{\alpha}_{t-1}\cdot\beta_t^2}{(1-\bar{\alpha}_t)^2}\cdot
    \E_{q(\x_t|\x_0)}\left[
    \left|\left|\hat\x_\btheta(\x_t, t)-\x_0\right|\right|_2^2\right] \, .
\end{equation}

Optimizing our neural network to minimise a  loss similar to \Cref{eq:loss_with_weighs}  will help us define a generative process that allocates as high a likelihood to the training data as possible. 
We will soon discuss how to use this to generate new samples that (approximately) are sampled from $p_\text{complex}$, but first we will summarise what we have found so far.

\subsubsection{Summary of the diffusion process and model training loss}
We will consider how to efficiently optimise this loss wrt.\ $\btheta$ in the following section, but let us first summarise where we are and how we got here:
\begin{itemize} 
\item 
Starting from a predetermined and fixed diffusion model $q(\x_t|\x_{t-1})$, we gradually diffuse a data-point $\x_0$ from the data distribution $p_\text{complex}$ into a point $\x_T$ from a simple distribution $p_\text{prior}$, typically assumed to be a standard Gaussian. 

\item 
We now aim to reverse the diffusion process, that is, find the ``inverse'' of  $q(\x_t|\x_{t-1})$, which we denote $p_\btheta(\x_{t-1}|\x_{t})$. 
We have the benefit of knowing where the reverse process is supposed to end up, i.e., we know that the marginal distribution for $t=0$,  $p_\btheta(\x_{0})$, must be equal to $p_\text{data}$, which again approximates $p_\text{complex}$.   

\item 
Starting out with the goal of maximising the likelihood 
$p_\btheta(\x_{0}) = \int_{\x_{1:T}} p(\x_T)\prod_{t=1}^T p_\btheta(\x_{t-1}|\x_{t}) \, d \x_{1:T}$, 
we ended up with the goal of minimising terms like
\begin{equation}
\E_{q(\x_{t}|\x_0)} \left[\DklP{q(\x_{t-1}|\x_t,\x_0)}{p_\btheta(\x_{t-1}|\x_t)}\right]\;.
\nonumber
\end{equation}
We argued that this is obtained by  choosing  $p_\btheta(\x_{t-1}|\x_t)$ to be a Gaussian with variance depending only on the scheduling parameters $\{\beta_t\}_{t=0}^T$. The mean, on the other hand, should be determined as a combination of $\x_t$ and $\x_0$, because we are starting from $\x_t$ and taking a step  intended to be in the direction of $\x_0$. 
Unfortunately, though, $\x_0$ is not known. Therefore, we create a neural network with weights $\btheta$ that uses the current value $\x_t$ and the time-index $t$ as inputs, and outputs an estimate for $\x_0$, denoted $\hat\x_\btheta(\x_t, t)$.
Loss terms like \Cref{eq:loss_with_weighs} help optimise $\btheta$.

\item 
Before continuing on to actually optimise the loss, consider first why one would expect that it is even possible to train a good model $\hat\x_\btheta(\x_t, t)$. 
Let us first consider $t=1$. Now, as the network is asked to generate $\hat\x_0 := \hat\x_\btheta(\x_1, t=1)$, it has two pieces of information: First, if $\x_1$ is meaningful, then the generated $\hat\x_0$ must be ``close'' to $\x_1$ -- because the pair $\left(\hat \x_0, \x_1\right)$ should be likely under the known model $q(\x_1|\x_0)$, and this process typically has a very small variance $(1-\alpha_1)\,\textbf{I}$. 
Second, $\hat\x_0$ should be likely under $p_\text{complex}$. For small values of $t$ one would thus expect to have a reasonable learning signal for learning $\hat\x_\btheta(\x_t, t)$. As $t$ grows larger, the strength of the signal decreases. 
While it is still true that the predicted $\hat\x_0$ should be likely under $p_\text{complex}$, the guiding from $\x_t$ decreases in $t$ (remember that the variance in $q({\x}_t | {\x}_0)$ is $(1-\Bar{\alpha}_t) \mathbf{I}$, see \Cref{eq:q_t_0}, and consider how quickly $\Bar{\alpha}_t$ drops in \Cref{fig:diffusion_parameters}). To accept that it still works, though, the argument can be made that if $\hat\x_{t-1} := \hat\x_\btheta(\x_{t}, t)$ is sufficiently precise for all $t\leq t'$ it is possible to utilise this for $t= t' +1$ too: $\hat\x_{t'}$ must be fairly close to $\x_{t'+1}$ to ensure that the continuation is likely under $q(\x_{t´+1}|\x_{t'})$ (notice how $\alpha_t$ drops off much slower than $\Bar{\alpha}_t$ in \Cref{fig:diffusion_parameters}). Step by step, this argument can be used to argue that it should be possible to learn $\hat\x_\btheta(\x_{t}, t)$ for all $t\leq T$.
\end{itemize}

\subsection{Reverse samplers}

Now that we have the general principles for defining the loss, we turn back to how to define the reverse diffusion process. 

\subsubsection{The DDPM sampler\label{sec:noise-driven-loss}}

The Denoising Diffusion Probabilistic Model (DDPM) \cite{ho_2020,nichol_2021}, learns the following parameterised Gaussian transitions:
\begin{equation}\label{eq:p_ddpm}
    p_\btheta(\x_{t-1}|\x_t) = \N(\x_{t-1}; \bmu_\btheta(\x_t, t), \Tilde\beta_t\,\mathbf{I}) \,,
\end{equation}
with $\Tilde\beta_t$ defined as in \Cref{eq:q_reverse_process}.
This leaves us with only having to learn $\bmu_\btheta(\x_t, t)$; compare this to \Cref{eq:p_t-1_t}. 
Furthermore, \cite{ho_2020} showed that rather than training a neural network to predict $\x_0$ (via $\tilde\x_\btheta(\x_t, t)$) using the loss we found in \Cref{eq:loss_with_weighs}, it is beneficial to take one more look at this and try to reformulate the objective. 
Starting from  \Cref{eq:xt-and-x0}, we can  represent $\x_0$ as
\begin{equation}
\label{eq:x0-via-eps}
\x_0 = \frac{1}{\sqrt{\bar{\alpha}_t}}\left(\x_t - \sqrt{1 - \bar{\alpha}_t}\beps_t\right)\, ,
\end{equation}
where $\beps_t$ is a multivariate standard Gaussian.

If we use the $\x_0$ from \Cref{eq:x0-via-eps} when calculating $\Tilde{\mu}(\x_t, \x_0)$ in  \Cref{eq:q_reverse_process}, we find that $\Tilde{\mu}(\x_t, \x_0)$, the expectation of the distribution $q(\x_{t-1}|\x_t,\x_0)$, and therefore  our desired choice for the expectation of $p_\btheta(\x_{t-1}|\x_t)$, too, can be re-written as 
\begin{equation}
\label{eq:mu-via-eps}
\Tilde{\mu}(\x_t, \x_0) = \frac{1}{\sqrt{\alpha_t}}\left(\x_t - \frac{1-\alpha_t}{\sqrt{1 - \bar{\alpha}_t}}\beps_t\right)\, .
\end{equation}
A main idea is now to use the representation \Cref{eq:mu-via-eps} for learning. This means, instead of learning the estimator $\hat\x_\btheta(\x_t,t)$, we will approximate $\beps_t$, and call the output from the learned model $\hat\beps_\btheta(\x_t,t)$.
Using this idea, we can (after some pencil-pushing) reformulate the KL-loss in \Cref{eq:KLnormal-simpler} as
\begin{equation} \label{eq:ddpm_loss_via_eps}
    \Loss_{t-1} = \frac{1}{2\Tilde{\beta}_t}
    \; \frac{(1-\alpha_t)^2}{\alpha_t(1-\bar{\alpha}_t)}
    \cdot
    \E_{q(\x_t|\x_0)}\left[
    \left|\left|\hat\beps_\btheta(\x_t, t)-\beps_t\right|\right|_2^2\right] \, .
\end{equation}
It was later shown empirically by \cite{ho_2020} that one obtained better results ignoring the weighing-term in the loss, so that
\begin{equation}\label{eq:simple_loss}
    \Loss_{t-1} \leftarrow
    \E_{q(\x_t|\x_0)}\left[
    \left|\left|\hat\beps_\btheta(\x_t, t)-\beps_t\right|\right|_2^2\right] \, .
\end{equation}
This gives us the training algorithm given in \Cref{algo:training}. 

\begin{algorithm}[h]
\begin{algorithmic}[1]
\Repeat
\State $\x_0\sim p_\text{data}$
\State $t \sim \text{Uniform}(\{1, \ldots, T\})$
\State $\beps \sim \N(\mathbf{0}, \mathbf{I})$
\State $\x_t \gets \sqrt{\Bar{\alpha}_t}\cdot\x_0 + \sqrt{1-\Bar{\alpha}_t} \cdot \beps$
\State $\btheta\gets \btheta - \eta \cdot\Grad \left|\left|\hat\beps_\btheta(\x_t, t)-\beps\right|\right|_2^2$
\Until{converged}
\end{algorithmic}
\caption{Training using learning-rate $\eta$}\label{algo:training}
\end{algorithm}

Given the DDPM's approach to finding $\beps_\btheta(\x_t,t)$, one step of the denoising process from $t$ to $t-1$ is done via
\begin{equation}\label{eq:ddpm_sample}
    \x_{t-1} = \frac{1}{\sqrt{\alpha_t}} \left( \x_t- \frac{1-\alpha_t}{\sqrt{1-\Bar{\alpha}_t}}\beps_\btheta(\x_t,t) \right) +\sqrt{\Tilde{\beta}_t} \z \,, \;\;\z\sim\N(\mathbf{0}, \mathbf{I}).
\end{equation}
Therefore, the reverse DDPM progresses as shown in \Cref{algo:ddpm_sample}.

\begin{algorithm}[h]
\begin{algorithmic}[1]
\State $\x_T \sim\N(\mathbf{0},\mathbf{I})$
\For {$t = T, \ldots,1$}
\If{$t>1$}
\State $\lambda\gets 1$ 
\Else
\State $\lambda\gets 0$ 
\EndIf
\State $\z \sim\N(\mathbf{0},\mathbf{I})$ 
\State $\x_{t-1} \gets \frac{1}{\sqrt{\alpha_t}} \left( \x_t- \frac{1-\alpha_t}{\sqrt{1-\Bar{\alpha}_t}}\cdot \beps_\btheta(\x_t,t) \right) +\lambda\cdot \sqrt{\Tilde{\beta}_t} \cdot\z$
\EndFor
\State\Return{$\x_0$}
\end{algorithmic}
\caption{Reverse DDPM}\label{algo:ddpm_sample}
\end{algorithm}

\begin{figure}
    \centering
        \includegraphics[width=.75\textwidth]{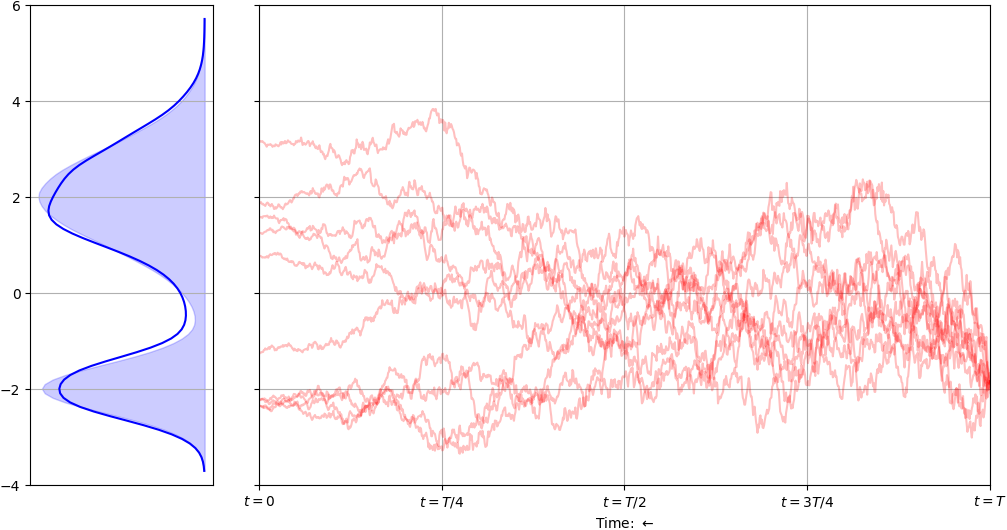} 
    \caption{Reverse process: Samples from DDPM, all starting from the same $\x_T$.  The reverse process starts from $t=T$ and moves towards $t=0$, so from right to left in the figure. The marginal drawn with a solid line at the far left shows the empirical distribution when repeating the samples several times, whereas the shaded area is the actual distribution $p_{\text{complex}}$ used to generate the training-data in this example. Notice how trajectories starting from a \textit{fixed} $\x_T$ still can cover the full distribution $p_{\text{complex}}$.
    }
    \label{fig:reverse_traces_DDPM}
\end{figure}

\Cref{fig:reverse_traces_DDPM} shows the DDPM-sampler at work for a simple univariate problem. The figure shows a number of trajectories all starting from the same sampled $\x_T$, and as the value of $t$ is reduced from $t=T$ to $t=0$ (i.e., moving from right to left) the end-result produces samples from a bimodal distribution quite similar to the target-distribution used to generate the training-data.  

\subsubsection{DDIM: Deterministic denoising}

Referring again to the DDPM sampler in \Cref{eq:ddpm_sample}, also depicted in \Cref{fig:reverse_traces_DDPM}, we see that the denoising procedure is stochastic, and we cannot determine which latent location $\x_T$ is the origin of a sampled $\x_0$. 
However, given the noise prediction model $\beps_\btheta(\x_t,t)$, we can alternatively create a \textit{deterministic} denoising process to create samples from the target distribution. 
This is called Denoising Diffusion Implicit Models (DDIMs), introduced in \cite{ddim_arxiv}.

A key observation is that the DDPM objective in \Cref{eq:loss_with_weighs} and \Cref{eq:simple_loss}, the latter repeated here for convenience, 
\begin{equation}\nonumber
    \Loss_{t-1} =
    \E_{q(\x_t|\x_0)}\left[
    \left|\left|\hat\beps_\btheta(\x_t, t)-\beps_t\right|\right|_2^2\right] \, ,
\end{equation}
only depends on $q(\x_t |\x_0)$ (this is the distribution used in the expectation), and not explicitly on the joint distribution $q(\x_{1:T} |\x_0)$ or each of  the noising-steps $q(\x_t |\x_{t-1})$. 
Since many joint distributions $q(\x_{1:T} |\x_0)$  can have the same marginals $q(\x_t |\x_0)$, \cite{ddim_arxiv} explore an alternative,  non-Markovian inference processes leading to new generative processes, but with the same surrogate objective function as DDPM: They consider a family of  distributions defined as
\begin{equation}
\nonumber
    q_\sigma (\x_{1:T} | \x_0) := q_\sigma (\x_T | \x_0) \prod^{T}_{t=2} q_\sigma(\x_{t-1} | \x_t,\x_0) \,  .
\end{equation}
Notice how the joint distribution $q(\x_{1:T} |\x_0)$ is defined in the direction of reversed time; where we previously had $q(\x_t |\x_{t-1})$ as our building-blocks, the joint of the forward process is now defined using the reverse view, and focus is on $q_\sigma(\x_{t-1} | \x_t,\x_0)$.
In order to arrive at an operational expression for the forward, i.e.\ the noising process, we can use Bayes' rule:
\begin{equation}\label{eq:q-xt-from-xt-1-non-markov}
    q_\sigma(\x_t | \x_{t-1},\x_0) = \frac{q_\sigma(\x_{t-1} | \x_t,\x_0) \cdot q_\sigma(\x_t|\x_0)}{q_\sigma(\x_{t-1}|\x_0)} \,.
\end{equation}
Notice that this forward process, $q_\sigma(\x_t | \x_{t-1},\x_0)$, is \textit{not} Markovian, as the distribution of $\x_t$ is conditioned on 
\textit{both} $\x_{t-1}$ \textit{and} $\x_0$. Please compare \Cref{eq:q-xt-from-xt-1-non-markov} to \Cref{eq:q-xt-from-xt-1} and note this important difference.

Next, \cite{ddim_arxiv} introduce the following functional representation\footnote{Note the subtle difference between our notation and what was used in \cite{ddim_arxiv}: Our $\bar{\alpha}_t$ corresponds to $\alpha_t$ in~\cite{ddim_arxiv}. We have chosen to use this notation to be consistent with \cite{ho_2020,nichol_2021}.}  of the denoising process for all $t>1$,
\begin{equation} \label{eq:q-xt-1-from-xt-non-markov}
    q_\sigma(\x_{t-1}|\x_t,\x_0) = 
    \N
    \left(
    \x_{t-1}; 
    \sqrt{\bar{\alpha}_{t-1}}\cdot \x_0 + \sqrt{1-\bar{\alpha}_{t-1}-\sigma^2_t} \cdot
    \frac{\x_t - \sqrt{\bar{\alpha}_t}\x_0}{\sqrt{1-\bar{\alpha}_t}},\sigma^2_t\cdot\mathbf{I}
    \right) \,  .
\end{equation}
This is selected to ensure that $q_\sigma(\x_t|\x_0)=\N(\sqrt{\bar{\alpha}_t}\x_0, (1-\bar{\alpha}_t)\mathbf{I})$ for all $t$, i.e.\ exactly our \Cref{eq:q_t_0}. 
The expression in \Cref{eq:q-xt-1-from-xt-non-markov} is similar to  \Cref{eq:q-xt-from-xt-1-and-x0-with-tildes}, but notice that the variance term has been replaced by the parameter $\sigma_t$ that can take any value, including zero.
Since $q_\sigma(\x_t|\x_0)=\N(\sqrt{\bar{\alpha}_t}\x_0, (1-\bar{\alpha}_t)\mathbf{I})$, just as is was for DDPM, and this term is the only one that played a role in the 
loss (see the expectation in  \Cref{eq:simple_loss}), we see that training for the two approaches will be identical: A neural network trained as a DDPM model
to estimate $\beps_\btheta(\x_t,t)$ can be directly used by the DDIM reverse process. 
This is really the exciting part: We do not have to train a different model for every choice of $\sigma_t$, but can use our DDPM objective to learn a generative process also for the \textit{non-Markovian forward processes} parameterised by $\sigma_t$!

The difference between DDPM and DDIM instead lies in how we define $p_\btheta(\x_{t-1}|\x_t)$, and thus how to ``translate'' the estimate of $\beps_\btheta(\x_t,t) $ into a new sample $\x_{t-1}$. The DDPM lead  us to investigate the distribution in  \Cref{eq:q-xt-from-xt-1-and-x0-with-tildes}, and obtaining the sampling-step in \Cref{eq:ddpm_sample}, repeated here for reference:
\begin{equation}
\nonumber
    \x_{t-1} = \frac{1}{\sqrt{\alpha_t}} \left( \x_t- \frac{1-\alpha_t}{\sqrt{1-\Bar{\alpha}_t}}\beps_\btheta(\x_t,t) \right) +\sqrt{\Tilde{\beta}_t}\cdot \z \,, \;\;\z\sim\N(\mathbf{0}, \mathbf{I}).
\end{equation}
Similarly, DDIM starts from \Cref{eq:q-xt-1-from-xt-non-markov}, and results in the sampling procedure
\begin{equation}\label{eq:ddim_sample}
    \x_{t-1} = 
    \underbrace{
    \frac{1}{\sqrt{\alpha_{t}}} 
    \left(\x_t - \sqrt{1-\bar{\alpha}_t} \cdot \beps_\btheta (\x_t, t) \right)}_{\text{Predicted } \x_0}
    + \underbrace{
    \sqrt{1-\bar{\alpha}_{t-1} - \sigma^2_t}\cdot \beps_\btheta (\x_t, t)}_{\text{Direction of } \x_t}
    + \underbrace{\sigma_t \cdot \z}_{\text{Noise}}\,, \;\;\z\sim\N(\mathbf{0}, \mathbf{I}).
\end{equation}
The DDIM-sampling of $\x_{t-1}$ consists of three parts: We first estimate $\x_0$ from $\beps_\btheta (\x_t, t)$, then (since $\x_{t-1}$ is not supposed to be $\x_0$ if $t>0$, but rather a step on a path between $\x_0$ and $\x_t$), a move is made in the direction from $\x_0$ towards $\x_t$. 
Finally, we add Gaussian noise, scaled by the constant $\sigma_t$. 
The last part, regarding the level of noise, gives us extra flexibility here as compared to the DDPM. For DDPM, noise had to be scheduled according to $\Tilde{\beta}_t$ as defined in \Cref{eq:q_reverse_process}. However, in DDIM,  $\sigma_t$ has been introduced as an extra constant that we can choose ourselves. 
If we were to select
\begin{equation}
\nonumber
\sigma_t = \sqrt{\frac{1-\bar{\alpha}_{t-1}}{1-\bar{\alpha}_t}} \sqrt{1-\frac{\bar{\alpha}_t}{\bar{\alpha}_{t-1}}}\, , 
\end{equation}
the forward process becomes Markovian and we re-obtain the DDPM generative process. On the other hand, it has become customary to rather select  $\sigma_t = 0$,  in which case the generative process is deterministic. This model is a so-called \textit{implicit} probabilistic model \cite{mohamed2017learning}. The generative process performed by such a model is coined a \textit{denoising diffusion implicit model}, short DDIM, by \cite{ddim_arxiv}. The important insight to remember is that the training objectives are equivalent for any value of $\sigma_t$, meaning that a model trained using the general DDPM process can be used for any generative process in the family, including DDIM, and thus be used to generate samples deterministically. This leads us to the sampling procedure in \Cref{algo:ddim_sample}.

\begin{algorithm}[h]
\begin{algorithmic}[1]
\State $\x_T \sim\N(\mathbf{0},\mathbf{I})$
\For {$t = T, \ldots,1$}
\If {$t>1$}
\State $\lambda\gets 1$
\Else
\State $\lambda\gets 0$
\EndIf
\State $
    \x_{t-1} \gets 
    \frac{1}{\sqrt{\alpha_t}} \;
     \left( \x_t-\sqrt{1-\Bar{\alpha}_t}\cdot\beps_\btheta(\x_t,t) \right) + \lambda\cdot \sqrt{1-\Bar{\alpha}_{t-1} } \cdot\beps_\btheta(\x_t,t) \,.
$
\EndFor
\State\Return{$\x_0$}
\end{algorithmic}
\caption{Reverse DDPM}\label{algo:ddim_sample}
\end{algorithm}

\begin{figure}
    \centering
        \includegraphics[width=.75\textwidth]{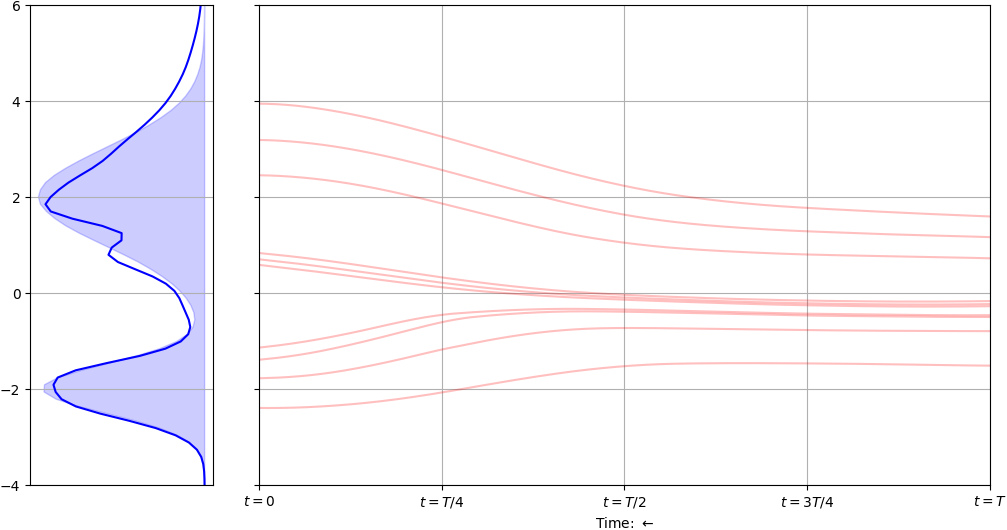}
    \caption{Reverse process using DDIM: $\x_T$ is sampled from a isotropic Gaussian distribution, while the following steps are deterministic. The reverse process is shown, hence moving from right to left. The solid-line marginal at the far left shows the empirical distribution when repeating the samples several times, whereas the shaded area is the actual distribution $p_{\text{complex}}$. Notice how the deterministic relationship between the starting $\x_T$ and the resulting $\x_0$ makes it necessary to sample many $\x_T$ to get a decent approximation to $p_{\text{complex}}$.}
    \label{fig:reverse_traces_DDIM}
\end{figure}

Having removed the stochastic element from the denoising process, this is fully deterministic, and we obtain the concept of a unique latent space for the model. To concretise: given the same initial sample $\x_T$, the DDIM process, \Cref{eq:ddim_sample}, always generates the same final sample, while the DDPM process, \Cref{eq:ddpm_sample}, generates different final samples for the same initial sample.
This can be observed in \Cref{fig:reverse_traces_DDIM}. 

The number of steps, $T$, in the forward, i.e.\ noising, process is an important hyper-parameter in diffusion models.  
As mentioned earlier, a small step size lets the denoising process approach a Gaussian, so that the generative process modelled with a Gaussian conditional distributions is a good approximation. 
This motivates large values of $T$, as in~\cite{ho_2020}.
However, in contrast to other generative models where the processes can be parallellised, all steps in the denoising process of DDPM must be performed sequentially. This gives the process of generation by denoising a significant disadvantage compared to other generative models.
An important result related to the developments of DDIM is therefore that one can in principle only evaluate the reverse process on a subset of points instead of doing it at each and every $t$, often without significant drops in quality of the generated objects. As this finding is not at the core of the definition of the model itself, we do not discuss the result here, but refer the interested reader to  \cite{ddim_arxiv}.

\section{Text-prompting}

\subsection{Text-conditioning}\label{sec:textcondition}
The most basic attempt at text-conditioning is described in \cite{Nichol2022glide}, and is tightly connected to the structure of the model: The U-Net \cite{ronneberger2015unet}, that is used for predicting $\hat\beps_\btheta(\x_t, t)$. The key is, for each text prompt $y$, to first encode the text into a sequence of $k$ tokens, and feed that into a transformer model.  The last layer of the transformer's token embeddings (a sequence of $k$ feature vectors) is concatenated to the attention context at each layer in the U-Net.
However, this approach reportedly leads to unstable image generation, and a need for stronger signal from the text-prompting. This is discussed next.

\subsection{Classifier guidance}
Text-conditioning (as presented in \Cref{sec:textcondition}) gives us the  opportunity to be gently steer the model towards objects that relate to the text-prompt. However, there is ample evidence showing that this can be improved even further. In this subsection we describe \textit{classifier guidance}, a way to guide the generation using an external classifier. We first describe the main idea and its mathematical foundation, then motivate how this can be used for text-guiding.

\subsubsection{The main idea}
Classifier guidance was introduced to diffusion models in \cite{dhariwal_nichol_arxiv}.
The core idea is inspired by work on GANs, but we will develop it independently of GAN-lingo here. Our starting-point is that we have learned the reverse (unconditional) process $p_\btheta(\x_{t-1}|\x_t)$, but now want to extend that using information from a separately trained probabilistic classifier. 
The classifier takes an input-point, e.g. an image $\x$, and produce a distribution over the random variable $Y$ that signifies the classes of the object (e.g., images can be of cars, horses, or people). This classifier is trained with  weights $\bxi$ different from the diffusion-model, so we will use $p_\bxi(y|\x)$ to signify the output of the classifier.  
Now, the idea is  to define the reverse diffusion $p_{\btheta,\,\bxi}(\x_t|\x_{t+1}, y)$. This means we are making the reverse \textit{conditional on the class}. This will make us say, e.g., ``generate an image of a horse'' instead of just asking for a general image (that could come from any class represented in the training data, of course including horses). 
\cite{dhariwal_nichol_arxiv} showed how this could be done using some simple tricks. First, assume that $Y \!\perp\!\!\!\perp \X_{t+1} | \X_t$. This means that any trace of the class variable that is present in $\X_{t+1}$ is also available after ``cleaning up'' to get $\X_t$. Therefore, if we want to guess on the class and have both $\x_t$ and $\x_{t+1}$ available, then using only the former will suffice. Then
\begin{alignat}{2}
p_{\btheta,\,\bxi} (\x_t|\x_{t+1}, y) 
    & = \frac{p_{\btheta,\,\bxi} (\x_t,\x_{t+1}, y) }{p_{\btheta,\,\bxi}(\x_{t+1}, y) } \nonumber\\
    & \propto  p_{\btheta,\,\bxi} (\x_t,\x_{t+1}, y)  \nonumber\\
    & =  p_{\btheta} (\x_{t+1}) \cdot p_{\btheta} (\x_t|\x_{t+1}) \cdot p_{\bxi} (y|\x_t) \nonumber\\
    & \propto p_{\btheta} (\x_t|\x_{t+1}) \cdot p_{\bxi} (y|\x_t) \, , \label{eq:classifier:introduced} 
\end{alignat}
where both proportionality statements follow from our interest on a distribution over only $\x_t$ so any distribution over the other variables are collected in proportionality constants. Further, the second equality follows from the conditional independence assumption: $p_{\btheta,\bxi} (y|\x_t, \x_{t+1}) = p_{\bxi} (y|\x_t)$. 
The consequence of \Cref{eq:classifier:introduced} is that we can utilise both the externally learned classifier $p_\bxi(y|\x)$ and the reverse diffusion process $p_{\btheta} (\x_t|\x_{t+1})$ 
as they are, and incorporate the classifier's output to guide the reverse diffusion towards creating objects seen as representatives from a particular class $y$.  
What is still problematic, though, is that we do not really know what distribution family $p_{\btheta,\,\bxi} (\x_t|\x_{t+1}, y)$ will belong to. 
Without knowing this, it will be difficult to sample from the distribution. 
$p_{\btheta} (\x_t|\x_{t+1})$ has been assumed to be a Gaussian, but we have no reason to make strong assumptions about the structure of the likelihood $p_{\bxi} (y|\x_t)$ (that is, as a function of $\x_t$, and not $y$). 
To simplify matters, we will use a one-step Taylor approximation of $\log p_{\bxi} (y|\x_t)$ around some value $\bnu$:
\begin{equation}
\nonumber
\log p_{\bxi} (y|\x_t) \approx \log p_{\bxi} (y|\x_t)_{\mid \x_t=\bnu} + \left(\x_t - \bnu\right)^\top \boldsymbol{\nabla}_{\!\x_t} \log p_{\bxi} (y|\x_t)_{\mid \x_t=\bnu}.
\end{equation}
Notice that since $\log p_{\bxi} (y|\x_t)_{\mid \x_t=\bnu}=\log p_{\bxi} (y|\bnu)$ is constant in $\x_t$, we will simplify an write that  
$\log p_{\bxi} (y|\x_t) \approx  \left(\x_t - \bnu\right)^\top \boldsymbol{\nabla}_{\!\x_t} \log p_{\bxi} (y|\x_t)_{\mid \x_t=\bnu} + C'$ where $C'$ collects terms not related to $\x_t$.
We can simplify further by utilising that also $\boldsymbol{\nabla}_{\!\x_t} \log p_{\bxi} (y|\x_t)_{\mid \x_t=\bnu}$ is constant in $\x_t$ after inserting that $\x_t=\bnu$,  and we get 
\begin{equation}
\label{eq:classifier:guide:contrbution}
\log p_{\bxi} (y|\x_t) \approx  \x_t^\top \boldsymbol{\nabla}_{\!\x_t} \log p_{\bxi} (y|\x_t)_{\mid \x_t=\bnu} + C.
\end{equation}

If we look back at \Cref{eq:classifier:introduced} and use \Cref{eq:classifier:guide:contrbution} to approximate the effect of the classifier, we see that we can express the classifier-guided reverse process  as 
\begin{alignat}{2}
\log p_{\btheta,\,\bxi} (\x_t|\x_{t+1}, y) & = \log p_{\btheta} (\x_t|\x_{t+1}) + \log p_{\bxi} (y|\x_t) + C_1 \nonumber\\
& \approx \log p_{\btheta} (\x_t|\x_{t+1}) + \x_t^\top \boldsymbol{\nabla}_{\!\x_t} \log p_{\bxi} (y|\x_t)_{\mid \x_t=\bnu} + C_2.
\label{eq:classifier:guide:combined}
\end{alignat}

Remember that we previously decided that $\log p_{\btheta} (\x_t|\x_{t+1}) $ should be a Gaussian with given mean and covariance (see \Cref{eq:p_t-1_t}). 
Let's call these parameters $\bmu_t$ and $\bSigma_t$ for simplicity. 
Further, note that a random variable $\Z$ follows the multivariate Gaussian with these parameters if and only if its log-density can be expressed as 
\begin{equation}
\log p(\z|\bmu_t,\bSigma_t)=-\frac{1}{2} \z^\top\bSigma_t^{-1}\z + \z^\top \bSigma_t^{-1} \bmu_t + \text{constant}\,,
\label{eq:gaussian:natural-pdf}
\end{equation}
where the term ``constant'' refers to terms independent of $\z$. 
Using this representation for $ \log p_{\btheta} (\x_t|\x_{t+1})$ in \Cref{eq:classifier:guide:combined} and deciding to do the Taylor approximation around $\bnu=\bmu_t$, we get
\begin{alignat}{2}
\log p_{\btheta,\,\bxi} (\x_t|\x_{t+1}, y) 
& \approx \log p_{\btheta} (\x_t|\x_{t+1}) + \x_t^\top \boldsymbol{\nabla}_{\!\x_t} \log p_{\bxi} (y|\x_t)_{\mid \x_t=\bmu_t} + C_2 \nonumber\\
&=-\frac{1}{2} \x_t^\top\bSigma_t ^{-1}\x_t + \x_t^\top \bSigma_t^{-1} \bmu_t + \x_t^\top \boldsymbol{\nabla}_{\!\x_t} \log p_{\bxi} (y|\x_t)_{\mid \x_t=\bmu_t} + C\nonumber\\
&=-\frac{1}{2} \x_t^\top\bSigma_t^{-1}\x_t + \x_t^\top \bSigma_t^{-1} \bmu_t + \x_t^\top \bSigma_t^{-1}\bSigma_t\boldsymbol{\nabla}_{\!\x_t} \log p_{\bxi} (y|\x_t)_{\mid \x_t=\bmu_t} + C \nonumber\\
&=-\frac{1}{2} \x_t^\top\bSigma_t^{-1}\x_t + \x_t^\top \bSigma_t^{-1} \left[\bmu_t + \bSigma_t\boldsymbol{\nabla}_{\!\x_t} \log p_{\bxi} (y|\x_t)_{\mid \x_t=\bmu_t}\right] + C
\label{eq:classifier:guide:done}
\end{alignat}

Comparing \Cref{eq:classifier:guide:done} to \Cref{eq:gaussian:natural-pdf} we realise that $\log p_{\btheta,\,\bxi} (\x_t|\x_{t+1}, y)$ is (approximately) the log-pdf of a Gaussian with mean $\bmu_t + \bSigma_t\boldsymbol{\nabla}_{\!\x_t} \log p_{\bxi} (y|\x_t)_{\mid \x_t=\bmu_t}$ and covariance $\bSigma_t$. Therefore the classifier-corrected reverse diffusion process is also (approximately) Gaussian, with known parameters. 
This result is quite fascinating: At each point in time, the classifier-guided reverse will sample from a distribution which has its mean nudged a little bit off the mean that we would have had without the classifier. The offset is $\bSigma_t\;\boldsymbol{\nabla}_{\!\x_t} \log p_{\bxi} (y|\x_t)_{\mid \x_t=\bmu_t}$. As $\bSigma_t$ is simply a constant times the identity matrix, the contribution from the classifier guidance is in 
$\boldsymbol{\nabla}_{\!\x_t} \log p_{\bxi} (y|\x_t)_{\mid \x_t=\bmu_t}$. 
Remember that the vector $\boldsymbol{\nabla}_{\!\x} \log p_{\bxi} (y|\x)$ points in the direction (in $\x$-space) for which $\log p_{\bxi} (y|\x)$ increases the most from a starting-point $\x$. 
This means that when     starting from an object $\x$, the clever thing to do to make it more like ``something from class $y$'' is to make an adjustment in the direction of $\boldsymbol{\nabla}_{\!\x} \log p_{\bxi} (y|\x)$. 
This is also exactly what goes on in \Cref{eq:classifier:guide:done}. 
So, while the developments leading to \Cref{eq:classifier:guide:done} were based on careful examination of the independence structure (\Cref{eq:classifier:introduced}) and a first-order Taylor approximation to the classifier's likelihood (\Cref{eq:classifier:guide:contrbution}), the end-result is also  intuitive and easily acceptable. Finally, \cite{dhariwal_nichol_arxiv} argue that one may want more or less focus on the classifier guidance, 
and they therefore introduce a hyper-parameter $s$ to determine this part's importance. Their final result is thus
\begin{equation} \label{eq:classifier_guided_w_s}
p_{\btheta,\,\bxi} (\x_{t-1}|\x_{t}, y) \sim \N
\left(
\hat\bmu_\btheta(\x_t, t) + s\,\Tilde{\beta}_t\boldsymbol{\nabla}_{\!\x} \log p_{\bxi} (y|\x)_{\mid \x=\hat\bmu(\x_t, t)}; \Tilde{\beta}_t\mathbf{I}
\right)
\;.
\end{equation}


\subsubsection{Using classifier guidance for text}
The description above is general, in the sense that we only require a probabilistic classifier $p_{\bxi} (y|\x)$ and did not make hard requirements about how the classifier is learned, or what the classes actually mean. In this subsection we will briefly discuss how to use the classifier-guided setup for text prompting. Of course, if we let the classifier denote captions, and think that there for each image is one and only one caption that is correct, then we can in principle learn a classifier over this (extremely huge) set of classes, and use \Cref{eq:classifier_guided_w_s} to guide the object generation towards an image that is relevant for the desired text. This is the goal of \cite{Nichol2022glide}. However, to avoid the challenges of learning the classifier, the authors use CLIP \cite{Radford21CLIP}. 

Very briefly, CLIP utilised a data-set of 400 million (image, text)-pairs to learn two learn a multi-modal embedding space by jointly training an image encoder $\mathbf{f}_\text{img}(\cdot)$ and a text encoder $\mathbf{f}_\text{txt}(\cdot)$. These two models are learned so that the cosine similarity between an image $\x$ and caption $y$, $\mathbf{f}_\text{img}(\x)^\top \mathbf{f}_\text{txt}(y)$ is maximised if $y$ is the caption for the image $\x$. 

With this in hand, \cite{Nichol2022glide} propose to adapt the classifier-guidance in \Cref{eq:classifier_guided_w_s} using $\mathbf{f}_\text{img}(\x_t)^\top \mathbf{f}_\text{txt}(y)$ as a proxy for $\log p_{\bxi} (y|\x)$, giving the mean of the distribution $p_{\btheta,\,\bxi} (\x_{t-1}|\x_{t}, y)$ as 
\begin{equation}
\nonumber
\hat\bmu_\btheta(\x_t, t) + s\,\Tilde{\beta}_t\boldsymbol{\nabla}_{\!{\x}} \left(\mathbf{f}_\text{img}(\x)^\top \mathbf{f}_\text{txt}(y)\right)_{\mid \x=\x_t}.
\end{equation}
A final comment needed here is that the original clip model trained $\mathbf{f}_\text{img}(\cdot)$ on \textit{actual} images, while our use-case is to analyse \textit{noisy} images $\x_t$, $t>0$.  Naturally, \cite{Nichol2022glide} were  therefore able to improve their results by retraining the image embeddings using noisy images, i.e., learn an image embedder $\mathbf{f}_\text{img}(\x_t, t)$ that receives a noisy image ($\x_t$) and the noise level (indirectly, via $t$) and produces an embedding in the same multi-modal embedding space.

\subsection{Classifier-free guidance}
In an attempt to do text-guiding without requiring  the ``external'' classifier $p_{\bxi} (y|\x)$ (or the CLIP embedding models 
$\mathbf{f}_\text{img}(\cdot)$ and $\mathbf{f}_\text{txt}(\cdot)$), \cite{ho2021classifierfree} proposed a way to train a diffusion-model without the reliance on external models. 

The starting-point is the estimator for $\hat\beps_\btheta(\x_t, t)$ as learned in \Cref{eq:ddpm_loss_via_eps}.
The idea is now to instead learn a \textit{conditional model}, $\hat\beps_\btheta(\x_t, y, t)$, that now also receives an image caption (or in practice a semantic embedding of the captions; it has been shown by \cite{saharia2022photorealistic} that pre-trained language-model can be used to embed the textual content without loss in image generation quality) as its an additional input $\y$. This means that the model is learned from (image, text)-pairs. Every now and then, i.e., randomly and with a fixed probability, the textual information is suppressed in the input, 
leading the model to also learn $\hat\beps_\btheta(\x_t, \y\leftarrow\emptyset, t)$. 

At the time of image generation, the system starts by generating the embedding of the prompt, resulting in the embedding $\y$. 
Then, the guided $\beps$-estimator is defined as
\begin{equation}
\nonumber
\Tilde\beps_\btheta(\x_t, \y, t) = 
\hat{\beps}_\btheta(\x_t, \y, t)
+ s\cdot \left(\hat\beps_\btheta(\x_t, \y, t) - \hat\beps_\btheta(\x_t, \emptyset, t)\right),
\end{equation}
where  $s$ is a hyper-parameter used to determine the strength of the guiding. 
\cite{ho2021classifierfree} show that if the $\hat\beps$-model is exact, then 
\begin{equation}
\label{eq:class_free_eps}
\boldsymbol{\nabla}_{\!\x_t} \log p(\y|\x_t)\propto 
\hat\beps_\btheta(\x_t, \y, t) - \hat\beps_\btheta(\x_t, \emptyset, t).    
\end{equation}
Looking at the terms in \Cref{eq:class_free_eps}, we can thus see that the idea is to approximate the gradient of the score function by
$\hat\beps_\btheta(\x_t, \y, t) - \hat\beps_\btheta(\x_t, \emptyset, t)$, 
and add a correction to the estimated $\hat{\beps}_\btheta(\x_t, \y, t)$  in that direction. 
Finally, $\Tilde\beps_\btheta(\x_t, \y, t)$ can now take the role of $\hat{\beps}_\btheta(\x_t, t)$ in the reversed sampler (being either DDMP and DDIM).

\section{Current trends}\label{sec:trends}
We end these lecture notes by making some quick comments about directions of ongoing research, with examples of relevant works. 
A major trend in the use of diffusion models at the time of writing is to further improve text-based image-generation, also extending to both 3D synthesis       \cite{Poole22Diffusion3D} as well as generation of video \cite{Xing2023DiffVideoSurvey}.  
Diffusion models are also used to generate structured datatypes, like text sequences \cite{Gong23DiffuseSequence} and protein folding \cite{Wu22ProteinFolding}, and has also been used for solving reinforcement learning problems  \cite{Ajay2023DecisionDefuser}. 
Furthermore, there is an ongoing development of new theoretical results. One interesting strand of work is related to  Bayesian Flow Networks \cite{Graves23BFN}, that are able to generate samples both form from  continuous \textit{and discrete} variables. 
	
\appendix
\section{Mathematical tricks}

\subsection{Monte Carlo estimation}\label{sec:mc_estimator}
In this subsection we will briefly discuss Monte Carlo methods for estimating an expectation via sampling. This is an often used trick to get away from calculating difficult integrals to find expected values. 
Assume we have a function $f_{\btheta}(\x)$ for which we want to calculate the expected value, over a random variable $\X$ following some distribution $p_\btheta$. 
When the required calculations cannot be done analytically, the standard approach is to use a \textit{sample average}, also known as the Monte Carlo (or simply MC) estimator:
\begin{equation}
\mathbb{E}_{\X\sim p_\btheta}\left[\,f_{\btheta}(\X)\,\right] = 
\int_\x  f_{\btheta}(\x)\cdot p_\btheta(\x) \, {\rm d}\x \approx \frac{1}{M} \sum_{m=1}^M f_{\btheta}\left(\x_{(m)}\right),
\label{eq:mean_as_avg}
\end{equation}
\noindent
where $\{\x_{(1)},\ldots, \x_{(M)}\}$ are $M$ independent samples we have drawn from the distribution $p_\btheta$.
This estimator is powerful because we only need to be able to generate the samples $\{\x_{(1)},\ldots, \x_{(M)}\}$, and we can use the MC technique almost\footnote{None of the functions considered here are problematic, so we will not discuss the assumptions underlying the MC estimator.}
without  any assumptions about the function $f_\btheta$ or the distribution $p_\btheta$. 
Let us use 
$\hat\bmu_{M} = \frac{1}{M} \sum_{m=1}^M f_{\btheta}\left(\x_{(m)}\right)$
as a notational shortcut for the sample mean based on $M$ samples. 
Since this estimator is generated by sampling, it is a random variable. It can easily be shown that 
\begin{equation}
\nonumber
\mathbb{E}_{\X\sim p_\btheta}\left[\hat\bmu_{M}\right] = \mathbb{E}_{\X\sim p_\btheta}\left[\,f_{\btheta}(\X)\,\right], ~ \mathbb{V}_{\X\sim p_\btheta}\left[\hat\bmu_{M}\right] = 
\frac{1}{M} \mathbb{V}_{\X\sim p_\btheta}\left[\,f_{\btheta}(\X)\,\right].
\end{equation}
Hence, the Monte Carlo estimator is unbiased (meaning that it has the correct value in expectation). Furthermore, if the variance $\mathbb{V}_{\X\sim p_\btheta}\left[\,f_{\btheta}(\X)\,\right]$ is finite, it follows that the estimator's variance monotonically decreases towards zero when we increase the sampling effort (that is, as $M\rightarrow\infty$).

\subsection{The reparameterisation trick}
A key concept in probabilistic AI, which will be useful for evaluating the loss function when training our diffusion model, relies on the so-called \textit{reparameterisation trick} \cite{KingmaWelling2014}. 
\remove{
The starting point for understanding this very useful idea is the approximation of an expected value. 
Assume we have a function $f_{\btheta}(\x)$ for which we want to calculate the expected value, over a random variable $\X$ following some distribution $p_\btheta$. When the required calculations cannot be done analytically, the standard approach is to use a \textit{sample average}, also known as the Monte Carlo estimator:
\begin{equation}
\mathbb{E}_{\X\sim p_\btheta}\left[\,f_{\btheta}(\X)\,\right] = 
\int_\x  f_{\btheta}(\x)\cdot p_\btheta(\x) \, {\rm d}\x \approx \frac{1}{M} \sum_{m=1}^M f_{\btheta}\left(\x_{(m)}\right),
\label{eq:mean_as_avg}
\end{equation}
\noindent
where $\{\x_{(1)},\ldots, \x_{(M)}\}$ are $M$ independent samples drawn from $p_\btheta$.
This estimator is powerful because we only need to be able to generate the samples $\{\x_{(1)},\ldots, \x_{(M)}\}$, and we can use the MC technique almost\footnote{None of the functions considered here are problematic, so we will not discuss the assumptions underlying the MC estimator. 
} 
without  any assumptions about the function $f_\btheta$ or the distribution $p_\btheta$. 
}
The starting point for understanding this very useful idea is the approximation of an expected value, see \Cref{sec:mc_estimator}.
Next, assume that the expected value is part of a loss function and that we want to optimise the loss with respect to the parameters $\btheta$. We then need the gradient of the expectation we just approximated: 
\begin{equation}
\Grad \E_{\X\sim p_\btheta}\left[\,f_{\btheta}(\X)\,\right] = 
\Grad \int_\x p_\btheta(\x)\cdot f_{\btheta}(\x)\,  {\rm d}\x  =
 \underbrace{\int_\x   p_\btheta(\x) \cdot \Grad f_\btheta(\x)\, {\rm d}\x}_{\text{Suitable for MC}} +
 \underbrace{\int_\x   f_\btheta(\x) \cdot \Grad p_\btheta(\x)\, {\rm d}\x}_{\text{NOT suitable for MC}}  
 .
 \label{eq:mean_grad}
\end{equation} 
\noindent
Notice that while we can use the MC estimate from \Cref{eq:mean_as_avg} to approximate the first integral on the right-hand-side in \Cref{eq:mean_grad} by simply evaluating $\Grad f_{\btheta}\left(\x_{(m)}\right)$ for each sample $\x_{(m)}$, the MC estimate cannot be used to approximate the second integral. This is because neither $\Grad p_\btheta(\x)$ nor $f_{\btheta}(\x)$ will in general be probability densities.
The second integral is therefore not an expectation, and thus it makes no sense to approximate it by a sample average.

Here comes the trick: Assume that the distribution for $\X$ has a special form, namely that there exists a random variable $\Y$ following some distribution $p$, and a differentiable function $g(\y,\btheta)$ so that $g(\Y,\btheta)$ is distributed as $p_\btheta$.
This is a mouthful, but the requirement is that while $\Y\sim p$ comes from a distribution that is \textit{not} parameterised by $\btheta$, we can still find the function $g$ so that $g(\Y,\btheta)$ has the same distribution as $\X$, i.e., $g(\Y,\btheta)\sim p_\btheta$. This means that we can replace the expectation over $\X\sim p_\btheta$ with an expectation over $\Y\sim p$:

\begin{alignat}{1}
\Grad \mathbb{E}_{\X\sim p_\btheta}\left[\,f_{\btheta}(\X)\,\right] 
& =   
\Grad \mathbb{E}_{\Y\sim p}\left[\,f_{\btheta}\left(g(\Y,\btheta)\right)\,\right] \label{eq:variable_change} \\
& = 
\mathbb{E}_{\Y\sim p}\left[\,\Grad  f_{\btheta}\left(g(\Y,\btheta)\right)\,\right]  \label{eq:no_theta_in_y} \\
& \approx 
 \frac{1}{M} \sum_{m=1}^M \Grad  f_{\btheta}\left(g(\y_{(m)},\btheta)\,\right) \label{eq:reparam_done}
.
\end{alignat}
\noindent
Here, 
\Cref{eq:variable_change} is a simple change of variables, 
\Cref{eq:no_theta_in_y} holds because the distribution of $\Y$ does not depend on $\btheta$ and we can therefore interchange the expectation and the gradient, and 
\Cref{eq:reparam_done} simply uses Monte Carlo sampling to approximate the expectation in \Cref{eq:no_theta_in_y}, as we did in \Cref{eq:mean_as_avg}.
Note that we need to use the chain rule for derivatives to calculate 
$\Grad f_{\btheta}\left(g(\y_{(m)},\btheta)\,\right)$. 
In applications, we typically select a low value of $M$ for doing these calculations, sometimes as low as $M=10$ or even $M=1$, so that the approximation of the gradient can be calculated quickly. 

\paragraph{Example:}
To see how this works, consider a situation where $p_\btheta(x)=\N(\theta_1,\theta_2^2)$, and let $f(x)=x^2/2$ so that $f'(x)=x$. 
We are interested in $\Grad \mathbb{E}_{\X\sim p_\btheta}\left[\,f_{\btheta}(\X)\,\right]$. Due to the simple structure of this problem, we can in fact find the solution analytically: $\mathbb{E}_{\X\sim p_\btheta}\left[\,\frac{1}{2}\X^2\,\right] = \frac{1}{2} \cdot(\theta_1^2 + \theta_2^2)$, hence 
$\Grad \mathbb{E}_{\X\sim p_\btheta}\left[\,f_{\btheta}(\X)\,\right]=\btheta$. Nevertheless, let us also consider how the result can be obtained using the reparameterization trick.
Define $g(y, \btheta) = \theta_1 + \theta_2\cdot y$, and let $Y$ be a random variable with $p(y)=\N(0, 1)$. The properties of the Gaussian distribution ensures that $g(Y, \btheta) \sim \N(\theta_1, \theta_2^2)$, which is equal to $p_\btheta$. 
We can now use the reparameterisation trick to approximate $\Grad \mathbb{E}_{\X\sim p_\btheta}\left[\,f_{\btheta}(\X)\,\right]$. 
Explicitly, we find
\begin{equation}\nonumber
\Grad \mathbb{E}_{\X\sim p_\btheta}\left[\,f_{\btheta}(\X)\,\right] \approx
 \frac{1}{M} \sum_{m=1}^M \Grad f\left(\theta_1 + \theta_2\cdot y \right)= 
 \frac{1}{M} \sum_{m=1}^M
\begin{bmatrix}
    \theta_1 + \theta_2\cdot y_{(m)}  \\
    y_{(m)}  \cdot   \left(\theta_1 + \theta_2\cdot y_{(m)}\right) 
\end{bmatrix},
\end{equation}

\noindent
where we used that $f(\theta_1 + \theta_2\cdot y) = \frac{1}{2}\left(\theta_1 + \theta_2\cdot y\right)^2$, and therefore

\begin{equation}
\Grad f(\theta_1 + \theta_2\cdot y)=
\begin{bmatrix}
\frac{\partial}{\partial\theta_1} f(\theta_1 + \theta_2\cdot y) \\
\frac{\partial}{\partial\theta_2} f(\theta_1 + \theta_2\cdot y)
\end{bmatrix} =
\begin{bmatrix}
    2\cdot\frac{1}{2}\left(\theta_1 + \theta_2\cdot y\right) \cdot \frac{\partial}{\partial\theta_1} (\theta_1 + \theta_2\cdot y)\\
    2\cdot\frac{1}{2}\left(\theta_1 + \theta_2\cdot y\right) \cdot \frac{\partial}{\partial\theta_2} (\theta_1 + \theta_2\cdot y)
\end{bmatrix}=
\begin{bmatrix} 
\left(\theta_1 + \theta_2\cdot y\right)\cdot 1\\
 \left(\theta_1 + \theta_2\cdot y\right)\cdot y
\end{bmatrix}.
\nonumber
\end{equation}
Since $Y\sim\N(0, 1)$ we know that $\E[Y]=0$ and $\E[Y^2]=1$,  hence the reparameterization-trick will produce the right result in the limit:
\[
\frac{1}{M} \sum_{m=1}^M
\begin{bmatrix} 
\theta_1 + \theta_2\cdot y_{(m)}\\
\theta_1 \cdot y_{(m)}+ \theta_2 \cdot y_{(m)}^2
\end{bmatrix}
\xrightarrow{M\rightarrow\infty}
\begin{bmatrix} 
\theta_1 + \theta_2\cdot\E[Y]\\
\theta_1\cdot\E[Y]+\theta_2\cdot\E[Y^2]
\end{bmatrix}
=
\begin{bmatrix} 
\theta_1 + 0\cdot\theta_2\\
0\cdot\theta_1+ 1\cdot\theta_2
\end{bmatrix}
=
\begin{bmatrix} 
\theta_1\\
\theta_2
\end{bmatrix}
=\btheta
.
\]
In this simple example we could thus calculate the exact value, and found that $\Grad \mathbb{E}_{\X\sim p_\btheta}\left[\,f_{\btheta}(\X)\,\right] = \btheta$. 
We were  also able to show that the same  result was obtained by the reparameterisation trick in the limit when $M\rightarrow\infty$.
This concludes the demonstration.

\bibliographystyle{apalike}
\bibliography{bibliography-edit}

\begin{thebibliography}{}

\bibitem[Ajay et~al., 2023]{Ajay2023DecisionDefuser}
Ajay, A., Du, E., Gupta, A., Tenenbaum, J.~B., Jaakkola, T.~S., and Agrawal, P.
  (2023).
\newblock Is conditional generative modeling all you need for decision making?
\newblock In {\em The Eleventh International Conference on Learning
  Representations}.

\bibitem[Dhariwal and Nichol, 2021]{dhariwal_nichol_arxiv}
Dhariwal, P. and Nichol, A. (2021).
\newblock Diffusion models beat {GAN}s on image synthesis.
\newblock {\em ArXiv}, abs/.2105.05233.

\bibitem[Feller, 1949]{feller_1949}
Feller, W. (1949).
\newblock On the theory of stochastic processes, with particular reference to
  applications.
\newblock In {\em Proceedings of the [First] Berkeley Symposium on Mathematical
  Statistics and Probability}, volume~1, pages 403--433. University of
  California Press.

\bibitem[Gong et~al., 2023]{Gong23DiffuseSequence}
Gong, S., Li, M., Feng, J., Wu, Z., and Kong, L. (2023).
\newblock {DiffuSeq}: Sequence to sequence text generation with diffusion
  models.
\newblock {\em ArXiv}, abs/2210.08933.

\bibitem[Graves et~al., 2023]{Graves23BFN}
Graves, A., Srivastava, R.~K., Atkinson, T., and Gomez, F. (2023).
\newblock Bayesian flow networks.
\newblock {\em ArXiv}, abs/2308.07037.

\bibitem[Ho et~al., 2020]{ho_2020}
Ho, J., Jain, A., and Abbeel, P. (2020).
\newblock Denoising diffusion probabilistic models.
\newblock In Larochelle, H., Ranzato, M., Hadsell, R., Balcan, M., and Lin, H.,
  editors, {\em Advances in Neural Information Processing Systems}, volume~33,
  pages 6840--6851. Curran Associates, Inc.

\bibitem[Ho and Salimans, 2021]{ho2021classifierfree}
Ho, J. and Salimans, T. (2021).
\newblock Classifier-free diffusion guidance.
\newblock In {\em NeurIPS 2021 Workshop on Deep Generative Models and
  Downstream Applications}.

\bibitem[Kingma and Welling, 2014]{KingmaWelling2014}
Kingma, D.~P. and Welling, M. (2014).
\newblock Auto-encoding variational {B}ayes.
\newblock {\em ArXiv}, abs/1312.6114.

\bibitem[Mohamed and Lakshminarayanan, 2017]{mohamed2017learning}
Mohamed, S. and Lakshminarayanan, B. (2017).
\newblock Learning in implicit generative models.
\newblock {\em ArXiv}, abs/1610.03483.

\bibitem[Nichol et~al., 2022]{Nichol2022glide}
Nichol, A., Dhariwal, P., Ramesh, A., Shyam, P., Mishkin, P., McGrew, B.,
  Sutskever, I., and Chen, M. (2022).
\newblock {GLIDE}: {T}owards photorealistic image generation and editing with
  text-guided diffusion models.
\newblock {\em ArXiv}, abs/2112.10741.

\bibitem[Nichol and Dhariwal, 2021]{nichol_2021}
Nichol, A.~Q. and Dhariwal, P. (2021).
\newblock Improved denoising diffusion probabilistic models.
\newblock In Meila, M. and Zhang, T., editors, {\em Proceedings of the 38th
  International Conference on Machine Learning}, volume 139 of {\em Proceedings
  of Machine Learning Research}, pages 8162--8171. PMLR.

\bibitem[Poole et~al., 2022]{Poole22Diffusion3D}
Poole, B., Jain, A., Barron, J.~T., and Mildenhall, B. (2022).
\newblock {DreamFusion}: {T}ext-to-{3D} using {2D} diffusion.
\newblock {\em ArXiv}, 2209.14988.

\bibitem[Radford et~al., 2021]{Radford21CLIP}
Radford, A., Kim, J.~W., Hallacy, C., Ramesh, A., Goh, G., Agarwal, S., Sastry,
  G., Askell, A., Mishkin, P., Clark, J., Krueger, G., and Sutskever, I.
  (2021).
\newblock Learning transferable visual models from natural language
  supervision.
\newblock In {\em Proceedings of the 38th International Conference on Machine
  Learning}, volume 139 of {\em Proceedings of Machine Learning Research},
  pages 8748--8763. PMLR.

\bibitem[Ronneberger et~al., 2015]{ronneberger2015unet}
Ronneberger, O., Fischer, P., and Brox, T. (2015).
\newblock U-{N}et: {C}onvolutional networks for biomedical image segmentation.
\newblock {\em ArXiv}, abs/1505.04597.

\bibitem[Saharia et~al., 2022]{saharia2022photorealistic}
Saharia, C., Chan, W., Saxena, S., Li, L., Whang, J., Denton, E., Ghasemipour,
  S. K.~S., Gontijo-Lopes, R., Ayan, B.~K., Salimans, T., Ho, J., Fleet, D.~J.,
  and Norouzi, M. (2022).
\newblock Photorealistic text-to-image diffusion models with deep language
  understanding.
\newblock {\em ArXiv}, abs/2205.11487.

\bibitem[Sohl-Dickstein et~al., 2015]{pmlr-v37-sohl-dickstein15}
Sohl-Dickstein, J., Weiss, E., Maheswaranathan, N., and Ganguli, S. (2015).
\newblock Deep unsupervised learning using nonequilibrium thermodynamics.
\newblock In Bach, F. and Blei, D., editors, {\em Proceedings of the 32nd
  International Conference on Machine Learning}, volume~37 of {\em Proceedings
  of Machine Learning Research}, pages 2256--2265, Lille, France. PMLR.

\bibitem[Song et~al., 2020]{ddim_arxiv}
Song, J., Meng, C., and Ermon, S. (2020).
\newblock Denoising diffusion implicit models.
\newblock {\em ArXiv}, abs/2010.02502.

\bibitem[Wu et~al., 2022]{Wu22ProteinFolding}
Wu, K.~E., Yang, K.~K., van~den Berg, R., Zou, J.~Y., Lu, A.~X., and Amini,
  A.~P. (2022).
\newblock Protein structure generation via folding diffusion.
\newblock {\em ArXiv}, abs/2209.15611.

\bibitem[Xing et~al., 2023]{Xing2023DiffVideoSurvey}
Xing, Z., Feng, Q., Chen, H., Dai, Q., Hu, H., Xu, H., Wu, Z., and Jiang, Y.-G.
  (2023).
\newblock A survey on video diffusion models.
\newblock {\em ArXiv}, abs/2310.10647.

\end{thebibliography}

\end{document}